\documentclass[10pt,twocolumn,letterpaper]{article}
\usepackage[pagenumbers]{cvpr} 
\definecolor{cvprblue}{rgb}{0.21,0.49,0.74}
\usepackage[pagebackref,breaklinks,colorlinks,citecolor=cvprblue]{hyperref}

\usepackage{graphicx}
\usepackage{amsmath}
\usepackage{amssymb}
\usepackage{booktabs}

\usepackage{bm}
\usepackage{url}
\usepackage{pifont}
\usepackage{bbding}
\usepackage{comment}
\usepackage{caption}
\usepackage{multirow}
\usepackage{makecell}
\usepackage{amsfonts}
\usepackage{float}
\usepackage{multicol}
\usepackage{ragged2e}
\usepackage[misc]{ifsym}
\usepackage{pifont}
\usepackage{dsfont}
\usepackage[hang,flushmargin]{footmisc}
\usepackage{xcolor,colortbl}

\usepackage[accsupp]{axessibility} 
\usepackage{tabularx} 
\usepackage{array}
\newcommand{\PreserveBackslash}[1]{\let\temp=\\#1\let\\=\temp}
\newcolumntype{C}[1]{>{\PreserveBackslash\centering}p{#1}}
\newcolumntype{R}[1]{>{\PreserveBackslash\raggedleft}p{#1}}
\newcolumntype{L}[1]{>{\PreserveBackslash\raggedright}p{#1}}

\title{\textcolor{blue}{Advancing Myopia To Holism}: Fully Contrastive Language-Image Pre-training}

\author{
Haicheng Wang$^{1,2\dagger}$, 
Chen Ju$^{1\dagger}$\textsuperscript{\ding{41}}, 
Weixiong Lin$^{1,2}$, 
Shuai Xiao$^{1}$\textsuperscript{\ding{41}}, 
Mengting Chen$^{1}$, 
Yixuan Huang$^{1}$, \\
Chang Liu$^{1,2}$, 
Mingshuai Yao$^{1}$, 
Jinsong Lan$^{1}$, 
Ying Chen$^{1}$, 
Qingwen Liu$^{1}$, 
Yanfeng Wang$^{2}$
\\
$^1$ Alibaba Group \ \
$^2$ Shanghai Jiao Tong University\\
{\tt\small anakin\_skywalker@sjtu.edu.cn, cju.void@gmail.com, shuai.xsh@alibaba-inc.com} \\
\url{https://github.com/anakin-skywalker-Joseph/Holistic-CLIP}
}

\newcommand\blfootnote[1]{
\begingroup 
\renewcommand\thefootnote{}\footnote{#1}
\addtocounter{footnote}{-1}
\endgroup 
}

\begin{document}

\twocolumn[{
\renewcommand\twocolumn[1][]{#1}
\maketitle
\maketitle

\vspace{-0.9cm}
\begin{center}
\centering
\includegraphics[width=0.98\linewidth]{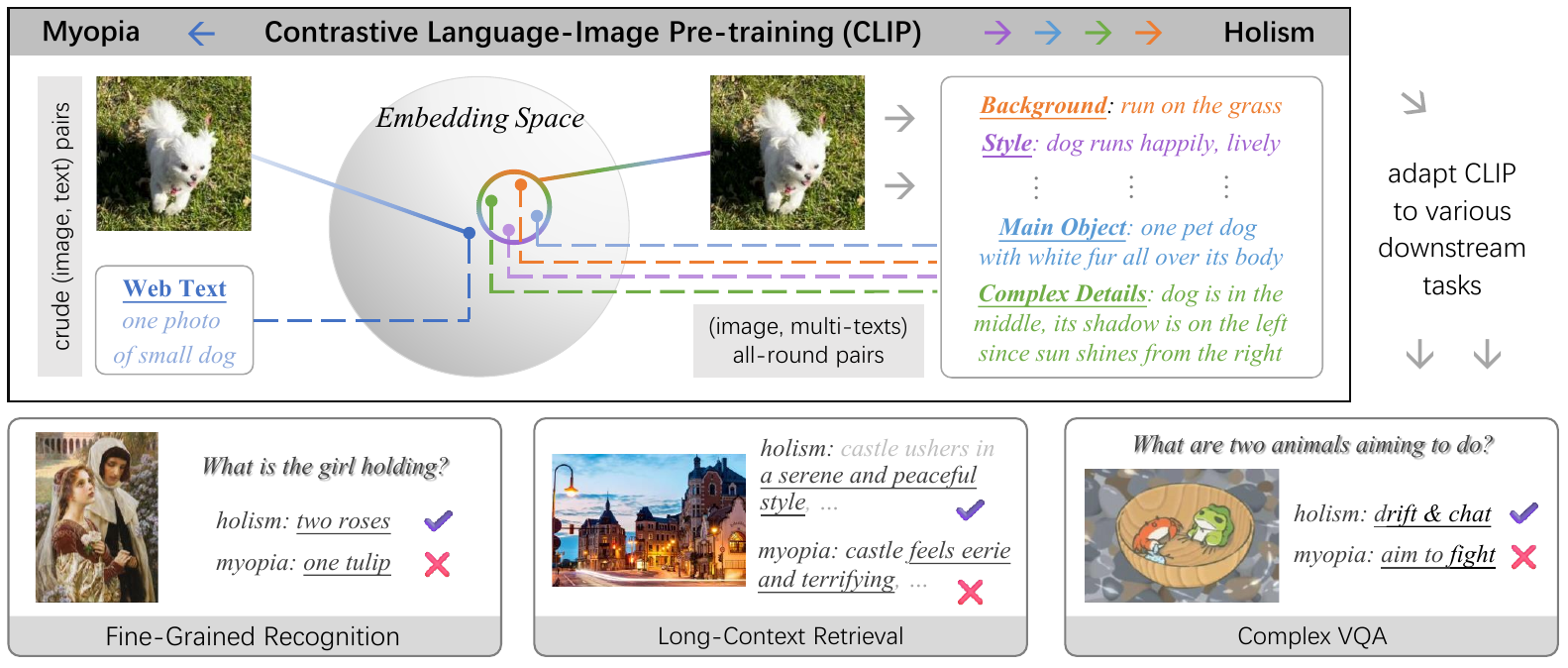}
\vspace{-0.3cm}
\captionof{figure}{\textbf{Myopia}. OpenAI's CLIP~\cite{clip} uses crude (image, text) web data for one-to-one contrastive alignment, causing serious myopia, {\em i.e.}, bias to monotonous short texts and shallow visual expressivity. \textbf{Holism}. We advance one holistic CLIP paradigm, by updating colorful (image, multi-texts) data from diverse views, levels; and designing multi-to-multi constrastive learning for image-text part-to-part matching.}
\vspace{0.4cm}
\label{figure:abs}
\end{center}
}]

\noindent \blfootnote{$\dagger$: These authors contribute equally to this work.} \noindent \blfootnote{\textsuperscript{\ding{41}}: Corresponding authors.}

\vspace{-0.2cm}
\begin{abstract}
In rapidly evolving field of vision-language models (VLMs), contrastive language-image pre-training (CLIP) has made significant strides, becoming foundation for various downstream tasks. However, relying on one-to-one (image, text) contrastive paradigm to learn alignment from large-scale messy web data, CLIP faces a serious myopic dilemma, resulting in biases towards monotonous short texts and shallow visual expressivity. To overcome these issues, this paper advances CLIP into one novel holistic paradigm, by updating both diverse data and alignment optimization. To obtain colorful data with low cost, we use image-to-text captioning to generate multi-texts for each image, from multiple perspectives, granularities, and hierarchies. Two gadgets are proposed to encourage textual diversity. To match such (image, multi-texts) pairs, we modify the CLIP image encoder into multi-branch, and propose multi-to-multi contrastive optimization for image-text part-to-part matching. As a result, diverse visual embeddings are learned for each image, bringing good interpretability and generalization. Extensive experiments and ablations across over ten benchmarks indicate that our holistic CLIP significantly outperforms existing myopic CLIP, including image-text retrieval, open-vocabulary classification, and dense visual tasks. 
\end{abstract}

\vspace{-0.5cm}
\section{Introduction} \label{sec:intro}
Recently, inspired by the tremendous progress of large language models in NLP domains, vision-language large models (VLMs) in CV domains also flourish, further triggered a boom in downstream tasks: closed-set understanding, open-vocabulary learning, and AIGC generation~\cite{rombach2022high,wang2023towards}. Hence, as the VLMs' cornerstone, contrastive language-image pre-training (CLIP)~\cite{clip} bears significant responsibility.

OpenAI leads CLIP's prevailing solution and most later variants follow the precedent set, {\em i.e.}, to bridge cross-modal embeddings effectively, CLIP first crawls 400M image-text pairs supplemented by simple cleaning, then optimizes via noisy contrastive learning for one-to-one alignment (Fig.\,\ref{figure:abs}). The paradigm rapidly expands data by extremely low manpower, and trains large-scale models to achieve strong performance. However, it also inevitably leads to bias in over-fitting the web data: (1) \textit{One Sidedness}. Crude text format is uniform, mostly presenting brief overviews without complex relationships, and some even carry the noise. (2) \textit{Visual Damage}. A colorful image only matches one compact text, diverse vision is largely discarded, with only a glance being used. (3) \textit{Chaotic Semantics}. Web texts are rather complicated with multi-level semantics, thus brutally aggregating them to the same image embedding fundamentally affects the pre-training quality. As a result, overfitting these biases doomed trained CLIP to be {\textbf{myopic}}, {\em i.e.}, performing poorly in long-context understanding~\cite{zhang2025long}, fine-grained visual perception~\cite{bianchi2024clip,monsefi2024detailclip}, and complexity modeling~\cite{subramanyam2023crepe}.

\vspace{-0.05cm}
To jump out the myopic dilemma, we are motivated by one \textit{\underline{blind men and an elephant} fable: combining cognition of multiple myopic men should give the holism.} Concretely, image carries explosive information than text, {\em e.g.}, people tell colors, objects, styles, events from multi-levels, multi-views for an image using various texts of varying lengths. Grouping all texts form full ideas of colorful vision.

\vspace{-0.05cm}
We accordingly propose an advanced pipeline for \textbf{holistic} contrastive language-image pre-training, updating data governance, model architecture and optimization paradigm. Firstly, data construction is primarily crucial, {\em i.e.}, evolving from uni-view/level (image, text) pairs to multi-view/level (image, multi-texts) pairs. And naturally, we leverage powerful VLMs for image-to-text captioning to form multiple texts with a little manpower. To encourage diversity in captioned texts, specific implementations are divided into two groups: one is to use multiple VLMs from various training data or architectures, implicitly expecting text complementarity; another is to carefully design prompts with distinct spirits, {\em e.g.}, focus guide, physical or sensory, gaze or glance, explicitly scattering text attention. Secondly, architecture needs to move forward to match (image, multi-texts) pairs. A vanilla idea is to reuse myopic CLIP's image encoder and text encoder, simply by encoding each text separately. Although feasible, it owns issues because an image covers colorful elements that cannot be sufficiently wrapped up by one embedding vector, which can cause serious visual damage. To tackle these issues, we modify the CLIP's image encoder into multi-branch visual outputs, covering two parameter-efficient schemes, {\em i.e.}, one is to initialize some CLS tokens to output corresponding embeddings; another is to extend the MLP of last layers to some parallel parts. By this way, multiple visual embeddings are gained in one forward. Finally, optimization paradigm for such holistic CLIP also requires evolution. Different from one-to-multi-positive constrastive learning used by recent studies~\cite{fan2024improving,zheng2025dreamlip}, we here design novel multi-to-multi constrastive learning by comprehensive image-text part-to-part matching. As the result, cross-modal alignment is achieved at multiple perspectives, granularities, and hierarchies. An additional benefit is that, diverse visual embeddings learned for each image, actually form the good semantic decomposition, allowing us to customize (select and combine several embeddings) for different downstream tasks, {\em i.e.}, generalizable and robust.

On over ten datasets, we extensively experiment to prove the generic superiority of holistic CLIP, across image-align-text understanding, image-to-text generation. Mostly, holistic CLIP greatly outperforms myopic CLIP, sometimes by 20\%. Thorough ablations are done to reveal each component's effectiveness, both quantitatively and qualitatively.

Our main contributions are summarized as follows:

$\bullet$ We pioneer to advancing contrastive language-image pre-training, from popular myopia to novel holism, by upgrading diverse data, model, and alignment paradigm.

$\bullet$ We develop one holistic CLIP pipeline, covering fully aligned data, multi-to-multi contrastive optimization. Colorful prompt spirits are designed to steer VLMs for multi-view/level image-text pairs. Exact matching of part-to-part vision-language are cogitated for training alignment. 

$\bullet$ We carry out numerous experiments to show superior gains of holism paradigm over prevalent myopic paradigm. Extensive ablations are studied for key components.

\section{Related Work}   \label{sec:related work}
\noindent \textbf{Vision-Language Pre-training (VLP)}~\cite{Jia21,zhai2022lit} uses large-scale web data for cross-modal alignment: CLIP~\cite{clip}, Florence~\cite{yuan2021florence}, FILIP~\cite{yao2021filip}, VideoCLIP~\cite{xu2021videoclip} and so on. In architectures, VLPs are mainly divided into: single tower~\cite{chen2020uniter,kim2021vilt}, twin towers~\cite{clip,Jia21}, and bridge tower~\cite{li2023blip,li2022blip,zhu2023minigpt}. For optimization, image-text one-to-one contrastive learning~\cite{hoffmann2022ranking,he2020momentum,cheng2023vindlu,li2022blip} is the mainstream, covering self supervision~\cite{liu2022exploiting,liu2024annotation}, weak supervision~\cite{li2023blip,cheng2023mixer} and partial supervision~\cite{ju2023constraint,liu2024audio}. As the foundation of VLMs, VLPs benefit many downstream tasks~\cite{ma2023diffusionseg,chen2024wear,ho2020denoising,rombach2022high,ramesh2022hierarchical,shi2020improving}.

But, excessive bias in web data leads to severe myopia in VLPs, showing shallow visual expressivity and diminished capacity for complex long-contexts. Recent work~\cite{fan2024improving,zhang2025long,zheng2025dreamlip,lai2025veclip} make minor improvements, by simply adding hard-sample data to continue training. To tackle these issues, we fully advance CLIP from myopia to holism, by upgrading diverse data, model encoder, and alignment paradigm.

\begin{figure*}[t]
\begin{center}
\vspace{-0.6cm}
\includegraphics[width=1\textwidth] {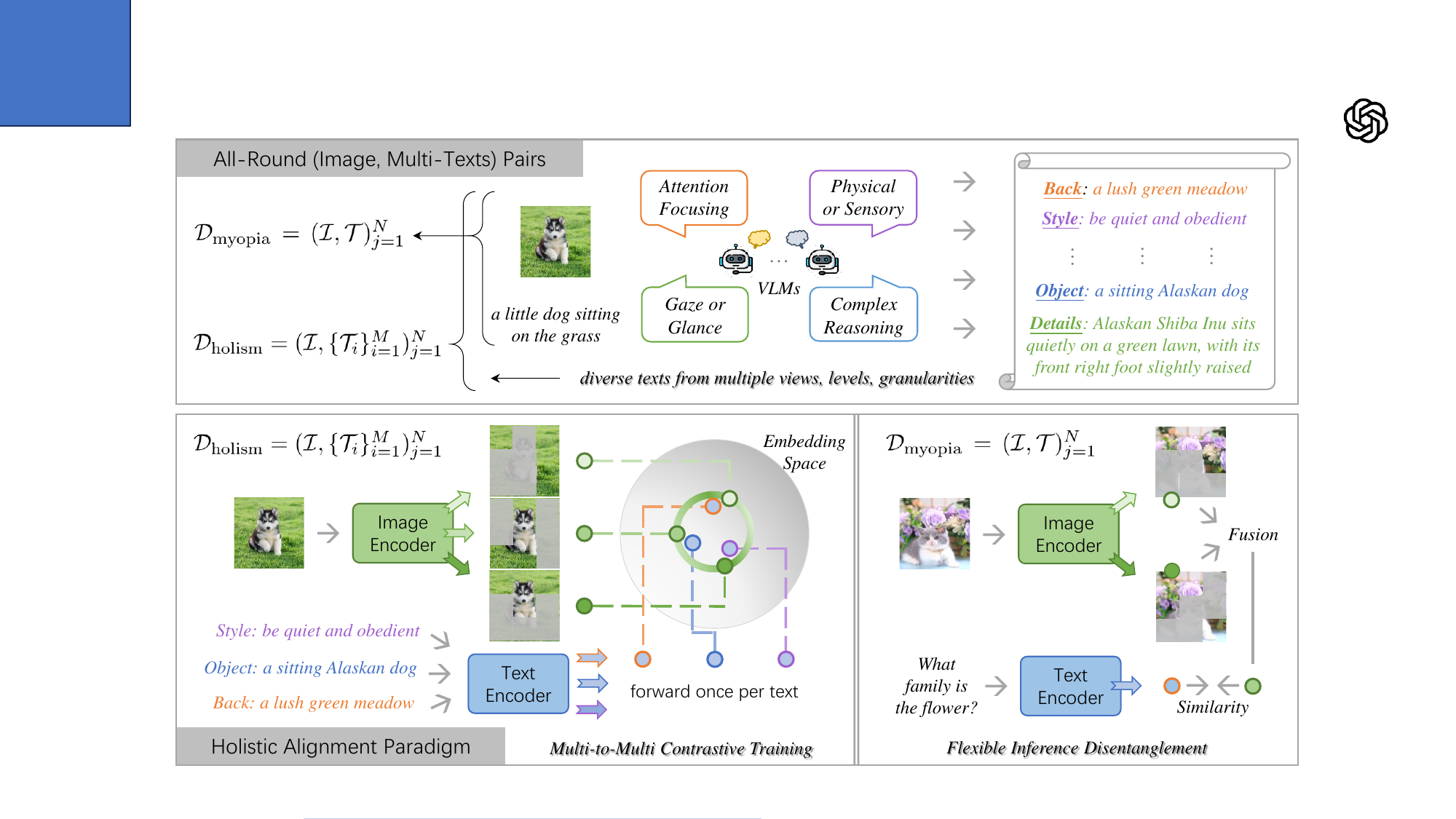}
\end{center}
\vspace{-0.5cm}
\caption{\textbf{Pipeline Overview of Holistic CLIP.} To evolve data from monotonous (image, text) pairs to colorful (image, multi-texts) pairs, we use powerful VLMs for captioning from multiple views, levels, and granularities. Diverse prompts are defined to encourage diversity. We then modify the CLIP image encoder into multi-branch, and optimize by multi-to-multi constrastive learning for part-to-part matching. During inference, flexible embedding customizations are available for different tasks, showing good interpretability and generalization.}
\label{fig:framework}
\end{figure*}

\vspace{0.1cm}
\noindent \textbf{Closed-set Alignment Understanding} has benefited many promising potentials from VLPs, in terms of cross-modal learning. Concretely, for image domains, object classification~\cite{zhou2022learning,ju2023turbo,ju2025turbo}, image-text retrieval~\cite{song2022clip,chen2023enhancing,cheng2023image}, segmentation~\cite{ye2021unsupervised}, and caption~\cite{mokady2021clipcap,luo2022clip4clip} all employ CLIP to improve performance. When it comes to video domains~\cite{zhao2020bottom,ju2022adaptive,ju2020point}, action recognition~\cite{ju2022prompting,wang2021actionclip}, grounding~\cite{ju2023constraint,liu2022exploiting,liu2023audio}, and temporal localization~\cite{ju2023distilling,ju2021divide} also gain solid progress.

\vspace{0.1cm}
{\noindent \bf Open-Vocabulary Learning} considers more realistic scenarios: generalize towards semantics that are unseen (zero-shot~\cite{wang2021actionclip,zhou2023non,wang2023towards}) or adopting several support samples (few-shot~\cite{ju2022prompting,yang2024multi,ju2023multi,ma2024open}). It describes concepts by free texts, then employ powerful VLPs to match text descriptions and visual samples. In the image domains, object classification, detection, and segmentation are well explored~\cite{du2022learning,ma2023open,yang2023multi,zhou2022learning}; while for video domains, recognition and localization of actions are mainly studied~\cite{huang2024froster,zhang2022ow,bao2022opental,cheng2024denoiser}.

\section{Method} \label{sec:method}
We advance contrastive language-image pre-trainings, from myopia to holism. Sec~\ref{rethink} reviews our motivation for holistic CLIP; Sec~\ref{com-data} reports image-text holistic alignment data, covering various views, levels and granularities; Sec~\ref{com-data} details multi-to-multi contrast for pipeline optimization.

\subsection{Motivation \& Overview}  \label{rethink}
Due to overfitting bias from web data, prevailing contrastive language-image pre-training is revealed to be myopic, {\em i.e.}, proficient in aligning short-text and main vision, but unable to understand complex, long or fine cases~\cite{bianchi2024clip,monsefi2024detailclip,zhang2025long}.

To advance myopia to holism, we follow the plain fable of \textit{blind men and an elephant}. In terms of information, one image is worth a thousand words. Regarding one image as an elephant, its paired text refers to the perception of the blind. When there are many blind men, these perceptions are multi-level, multi-granular, and multi-angle. One full-scale alignment can be formed by grouping all perceptions. As shown in Figure~\ref{fig:framework}, we propose novel holistic CLIP, an advanced pipeline updating data governance, model architecture and optimization paradigm.

\subsection{All-Round Vision-Language Data} \label{com-data}
To optimize contrastive language–image pre-training holistically, the data construction of multi-view/level alignment is primarily crucial. Hence, given the uni-view/level dataset $\mathcal{D}_{\mathrm{myopia}}=(\mathcal{I}, \mathcal{T})_{j=1}^N$ collected from web, our propose is to describe diverse texts $\{\mathcal{T}_i\}_{i=1}^M$ for each image, which could somewhat complement each other and work together to obtain comprehensive information. The resultant (image, $M$-texts) pairs form dataset $\mathcal{D}_{\mathrm{holism}} = (\mathcal{I}, \{\mathcal{T}_i\}_{i=1}^M)_{j=1}^N$.

Due to the enormous size of descriptive texts, the labors are usually unaffordable. Motivated by the impressive performance, we naturally seek vision-language large models (VLMs) $\Psi^{\mathrm{VLM}}$ for help, {\em i.e.}, synthesize descriptive texts for images, from various perspectives/granularities.

For specific implementations, one common idea~\cite{liu2023mllms,zheng2025dreamlip,lai2025veclip} is to utilize $M$-VLMs for re-captioning, expecting that VLMs from various data or architectures should implicitly complement each other's information. Formally, 
\begin{align} \label{eq:1}
\{\mathcal{T}_i\}_{i=1}^M = \{\Psi^{\mathrm{VLM}}_{i}(\mathcal{I}, \mathcal{P})\}_{i=1}^M, 
\end{align}
where $\mathcal{P}$ refers to prompts. This idea implicitly constrains prompt diversity by leveraging differences within $M$-VLMs themselves. Yet, it might  face issues in practice: deploying multi-VLMs is highly complex. We thus take another pragmatic path to ensure feasibility, {\em i.e.}, carefully designing diverse prompts to explicitly drive a single VLM. Specifically, prompts are responsible for captioning images in terms of breadth and depth, under four distinct spirits:

\underline{\textit{Focus Guide}} indicates whether attention should be laid on foreground objects or general background substances. \underline{\textit{Physical or Sensory}} refers to describe solid nouns or feeling styles. \underline{\textit{Gaze or Glance}} define captions are dense long-details or compact short-overview. \underline{\textit{Complex Reasoning}} distinguishes relationship or sequence for entities.

With above spirits, we could write $M$ prompts $\{\mathcal{P}_{i}\}_{i=1}^M$, use a representative VLM~\cite{chen2024internvl} to output $M$ colorful texts.
\begin{align} \label{eq:2}
\{\mathcal{T}_i\}_{i=1}^M = \Psi^{\mathrm{VLM}}(\mathcal{I}, \{\mathcal{P}_{i}\}_{i=1}^M). 
\end{align}
These captioned texts, along with the raw texts of $\mathcal{D}_{\mathrm{myopia}}$, constitute the ultimate data $\mathcal{D}_{\mathrm{holism}}$.

\vspace{0.1cm}
\noindent \textbf{Remark}. In supplementary materials, we offer many qualitative cases of the defined prompts and captioned texts. For each image, we quantitatively calculate the similarity~\cite{thakur-2020-AugSBERT} between multi-texts obtained by Eq.\,(\ref{eq:1}) vs Eq.\,(\ref{eq:2}). Statistically, their mean/variance are $0.58/0.043$ vs $0.48/0.052$, showing text diversity is pursued and colorful prompts are more effective. Table~\ref{tab:ablation1} also ablates for the text number. Besides, as web data is intricate, $\mathcal{D}_{\mathrm{myopia}}$ carries noisy texts, building $\mathcal{D}_{\mathrm{holism}}$ also finishes denoising by the way.

\subsection{Holistic Alignment Optimization}
We here take steps towards fully contrastive pre-training. 

\subsubsection{Vanilla CLIP (O2O)}
Vanilla CLIP~\cite{clip} deals with plain (one-image to one-text) pairs $\mathcal{D}_{\mathrm{myopia}}=(\mathcal{I}, \mathcal{T})_{j=1}^N$, hence it employs two separate encoders $\Phi^{\mathrm{img}}$/$\Phi^{\mathrm{txt}}$ for modality embeddings, which can be referred to:
$\mathbf{v} = \Phi^{\mathrm{img}}(\mathcal{I}), \,
\mathbf{t} = \Phi^{\mathrm{txt}}(\mathcal{T})$.

The optimization loss is defined as the average of two symmetric InfoNCE losses $(\mathcal{L}^{\mathrm{T2I}} + \mathcal{L}^{\mathrm{I2T}})/2$, which is actually \underline{one-to-one contrastive learning} between text-image: 
\begin{equation}
\begin{aligned}
\mathcal{L}^{\mathrm{T2I}}_{\mathrm{O2O}} = - \sum_{j=1}^K \log \frac{\exp(\langle \mathbf{v}_j, \mathbf{t}_j \rangle / \tau)}{\sum_{k=1}^K \exp(\langle \mathbf{v}_k, \mathbf{t}_j \rangle / \tau)}, 
\\
\mathcal{L}^{\mathrm{I2T}}_{\mathrm{O2O}} = - \sum_{j=1}^K \log \frac{\exp(\langle \mathbf{t}_j, \mathbf{v}_j \rangle / \tau)}{\sum_{k=1}^K \exp(\langle \mathbf{t}_k, \mathbf{v}_j \rangle / \tau)},
\end{aligned}
\end{equation}
where $\langle \cdot, \cdot \rangle$ denotes the cosine similarity, $\tau$ is one learnable temperature, and $K$ refers to the batch size.

Yet here, we are facing one holistic cross-modal dataset $\mathcal{D}_{\mathrm{holism}} = (\mathcal{I}, \{\mathcal{T}_i\}_{i=1}^M)_{j=1}^N$, whose elements are the (one-image to $M$-texts) pairs. Thus, the CLIP pipeline urgently needs improvements. One intuitive idea is to transform InfoNCE into multi-positive contrastive learning~\cite{fan2024improving,liu2023mllms,zheng2025dreamlip}.

\subsubsection{One-to-Multi Contrastive Learning (O2M)} 
For modal embeddings, we also formalize image and text encoders as $\Phi^{\mathrm{img}}$ and $\Phi^{\mathrm{txt}}$. Then naturally, the dual-modal embeddings can be calculated as follows:
\[
\mathbf{v} = \Phi^{\mathrm{img}}(\mathcal{I}), 
\quad \,
\{\mathbf{t}_i\}_{i=1}^M = \Phi^{\mathrm{txt}}(\{\mathcal{T}_i\}_{i=1}^M).
\]

For encoder optimization, a ($\mathbf{v}, \{\mathbf{t}_i\}_{i=1}^M$) pair is used to bring $M$ texts and single image closer together in the embedding space. Let's take text-to-image loss as an example, one-to-multi-positive constrastive learning is:
\begin{equation} 
\mathcal{L}^{\mathrm{T2I}}_{\text{O2M}} = - \sum_{j=1}^K \sum_{i=1}^M \log \frac{\exp(\langle \mathbf{v}_j, \mathbf{t}_{i,j} \rangle / \tau)}{\sum_{k=1}^K \exp(\langle \mathbf{v}_k, \mathbf{t}_{i,j} \rangle / \tau)},
\end{equation}

Although it sounds feasible, one-to-multi-positive contrastive learning still suffers from many issues: (1) \textit{Chaotic Semantics}. $M$ texts are from multi-view and multi-level, resulting in significant semantic differences. Brutally aggregating them at the same embedding position fundamentally affects the training procedure. (2) \textit{Visual Damage}. An image covers colorful elements, which cannot be sufficiently wrapped up by one embedding vector. During the damage of vision diversity, the resultant image embeddings become inexplicable, losing the capacity to decompose effective information for different downstream tasks.

\begin{table*}[t]
\centering
\footnotesize
\caption{\textbf{Elementary Short-Text Retrieval.} There are two solutions to get diverse texts of multi-level and multi-view: uni-VLM and multi-prompts, or multi-VLMs and uni-prompt. For multi-VLMs, we employ InternVL2~\cite{chen2024internvl}, Minigpt4~\cite{zhu2023minigpt}, LLaVA1.5~\cite{liu2024improved}, QwenVL2~\cite{bai2023qwen}, and set one unified prompt refer to~\cite{liu2023mllms}. For multi-prompts, we adopt InternVL2 and put prompt details in supplementary materials. For full comparisons under the same data, we also reproduce recent O2M methods~\cite{fan2024improving,zheng2025dreamlip,liu2023mllms} optimized by one-to-multi contrastive learning.}
\vspace{-0.3cm}
\begin{tabular}{cC{0.5CM}|C{1CM}|C{0.5CM}C{0.6CM}C{0.5CM}|C{0.4CM}C{0.4CM}C{0.6CM}C{0.4CM}C{0.4CM}C{0.7CM}|C{0.4CM}C{0.4CM}C{0.6CM}C{0.4CM}C{0.4CM}C{0.7CM}}
\toprule
\multicolumn{2}{c|}{\multirow{3}{*}{Method}} & \multirow{3}{*}{\begin{tabular}[c]{@{}c@{}}Training\\ Text\end{tabular}} & \multicolumn{3}{c|}{\multirow{2}{*}{Caption}} & \multicolumn{6}{c|}{MSCOCO} & \multicolumn{6}{c}{Flickr30} \\ \cline{7-18} 
\multicolumn{2}{c|}{} &  & \multicolumn{3}{c|}{} & \multicolumn{3}{c}{I2T} & \multicolumn{3}{c|}{T2I} & \multicolumn{3}{c}{I2T} & \multicolumn{3}{c}{T2I} \\ \cline{4-18} 
\multicolumn{2}{c|}{} &  & VLM & Prompt & Num & R@1 & R@5 & R@10 & R@1 & R@5 & R@10 & R@1 & R@5 & R@10 & R@1 & R@5 & R@10 \\ \hline \hline
\multicolumn{2}{c|}{CLIP~\cite{clip}} & CC3M & - & - & 1 & 13.6 & 32.9 & 43.9 & 13.4 & 32.6 & 44.1 & 30.8 & 56.2 & 68.1 & 31.9 & 58.0 & 68.3 \\ \hline
\multicolumn{2}{c|}{\multirow{3}{*}{O2M~\cite{fan2024improving,zheng2025dreamlip,liu2023mllms}}} & \multirow{3}{*}{\begin{tabular}[c]{@{}c@{}}CC3M\\+Caption\end{tabular}} & Uni & Uni & 2 & 20.0 & 42.9 & 54.3 & 20.1 & 42.9 & 53.8 & 50.5 & 74.4 & 83.2 & 49.1 & 74.7 & 82.6 \\
\multicolumn{2}{c|}{} &  & Uni & Multi & 5 & 24.5 & 48.6 & 59.9 & 26.3 & 49.4 & 60.9 & 60.7 & 81.3 & 88.1 & 61.7 & 83.2 & 88.8 \\
\multicolumn{2}{c|}{} &  & Multi & Uni & 5 & 27.1 & 51.8 & 64.3 & 27.9 & 53.8 & 64.5 & 63.9 & 85.3 & 88.9 & 63.8 & 84.4 & 89.7 \\ \hline 
\multicolumn{1}{c|}{\multirow{6}{*}{Ours}} & \multirow{3}{*}{$\Psi_{\mathrm{CLS}}$} & \multirow{3}{*}{\begin{tabular}[c]{@{}c@{}}CC3M\\+Caption\end{tabular}} & Uni & Uni & 2 & 21.6 & 44.9 & 55.6 & 20.9 & 44.8 & 55.4 & 51.0 & 74.6 & 82.0 & 51.0 & 74.9 & 83.9 \\
\multicolumn{1}{c|}{} &  &  & Uni & Multi & 5 & 28.0 & 52.4 & 63.4 & 27.8 & 53.1 & 64.1 & 62.9 & 85.1 & 89.9 & 64.2 & 85.7 & 91.1 \\
\multicolumn{1}{c|}{} &  &  & Multi & Uni & 5 & 30.0 & 54.5 & 65.8 & 30.2 & 54.7 & 66.2 & 66.0 & 86.1 & 91.3 & 65.9 & 87.3 & 91.9 \\ \cline{2-18} 
\multicolumn{1}{c|}{} & \multirow{3}{*}{$\Psi_{\mathrm{MLP}}$} & \multirow{3}{*}{\begin{tabular}[c]{@{}c@{}}CC3M\\+Caption\end{tabular}} & Uni & Uni & 2 & 20.1 & 43.7 & 55.0 & 20.6 & 43.4 & 55.1 & 50.9 & 74.8 & 82.8 & 51.2 & 75.7 & 83.8 \\
\multicolumn{1}{c|}{} &  &  & Uni & Multi & 5 & 28.2 & 53.0 & 64.4 & 27.4 & 52.4 & 64.0 & 63.7 & 85.3 & 90.6 & 63.9 & 85.8 & 90.5 \\
\multicolumn{1}{c|}{} &  &  & Multi & Uni & 5 & 31.2 & 55.7 & 66.5 & 30.7 & 55.9 & 66.9 & 66.5 & 86.4 & 91.2 & 66.4 & 86.9 & 92.2 \\ \hline \hline
\multicolumn{2}{c|}{CLIP~\cite{clip}} & CC12M & - & - & 1 & 25.2 & 49.4 & 61.9 & 25.3 & 50.0 & 61.7 & 56.9 & 81.6 & 87.4 & 55.2 & 81.9 & 87.6 \\  \hline 
\multicolumn{2}{c|}{\multirow{3}{*}{O2M~\cite{fan2024improving,zheng2025dreamlip,liu2023mllms}}} & \multirow{3}{*}{\begin{tabular}[c]{@{}c@{}}CC12M\\+Caption\end{tabular}} & Uni & Uni & 2 & 31.8 & 56.6 & 67.6 & 30.5 & 55.8 & 67.4 & 67.9 & 88.3 & 92.9 & 67.2 & 88.8 & 92.7 \\
\multicolumn{2}{c|}{} &  & Uni & Multi & 5 & 33.9 & 60.3 & 71.6 & 34.3 & 59.6 & 70.5 & 70.6 & 90.2 & 94.3 & 69.5 & 90.8 & 94.6 \\
\multicolumn{2}{c|}{} &  & Multi & Uni & 5 & 36.6 & 63.3 & 73.2 & 38.5 & 64.8 & 74.7 & 75.3 & 93.5 & 95.8 & 76.3 & 93.8 & 96.5 \\ \hline 
\multicolumn{1}{c|}{\multirow{6}{*}{Ours}} & \multirow{3}{*}{$\Psi_{\mathrm{CLS}}$} & \multirow{3}{*}{\begin{tabular}[c]{@{}c@{}}CC12M\\+Caption\end{tabular}} & Uni & Uni & 2 & 32.7 & 58.6 & 69.3 & 32.6 & 58.4 & 68.6 & 69.0 & 90.4 & 94.1 & 68.3 & 89.8 & 94.9 \\
\multicolumn{1}{c|}{} &  &  & Uni & Multi & 5 & 36.3 & 63.1 & 73.4 & 37.6 & 63.1 & 73.7 & 76.5 & 93.1 & 96.0 & 76.2 & 92.6 & 96.3 \\
\multicolumn{1}{c|}{} &  &  & Multi & Uni & 5 & 39.4 & 64.9 & 74.6 & 40.8 & 66.0 & 75.9 & 77.3 & 93.7 & 96.8 & 79.0 & 94.0 & 96.7 \\  \cline{2-18} 
\multicolumn{1}{c|}{} & \multirow{3}{*}{$\Psi_{\mathrm{MLP}}$} & \multirow{3}{*}{\begin{tabular}[c]{@{}c@{}}CC12M\\+Caption\end{tabular}} & Uni & Uni & 2 & 32.0 & 57.2 & 68.2 & 31.9 & 57.3 & 68.0 & 69.5 & 88.7 & 93.3 & 70.4 & 89.2 & 92.8 \\
\multicolumn{1}{c|}{} &  &  & Uni & Multi & 5 & 38.2 & 64.4 & 74.1 & 38.2 & 63.5 & 73.8 & 76.6 & 93.0 & 95.6 & 76.5 & 93.4 & 95.6 \\
\multicolumn{1}{c|}{} &  &  & Multi & Uni & 5 & 39.2 & 65.0 & 74.3 & 40.2 & 65.8 & 75.7 & 77.7 & 93.6 & 96.6 & 79.2 & 94.2 & 96.8 \\ \bottomrule
\end{tabular}
\label{tab:shorttext retrieval}
\end{table*}

\subsubsection{Multi-to-Multi Contrastive Learning (M2M)}
To solve the above issues, we are inspired by one MoE intuition~\cite{du2022glam} from LLM sparsity: \textit{if there is corresponding text for induction, the image encoder could output differentiated expert branches to decompose diversity, achieving accurate part-to-part alignment without losing information.}

In terms of network architecture, we modify the classic image encoder $\Phi^{\mathrm{img}}$ to enable the output of multi-branch visual embeddings. Concretely, two schemes are leveraged based on the $\Phi^{\mathrm{img}}$ structure. (1) \underline{$\Psi_{\mathrm{CLS}}$}. We initialize the $H$ CLS tokens to output $H$ different embeddings. (2) \underline{$\Psi_{\mathrm{MLP}}$}. We extend the MLP of last three layers to parallel $H$ parts. To reduce appended parameters, we only apply this modification to the second FFN in MLP blocks. Finally, in one forward, $H$ visual embeddings are obtained. Formally, 
\[
\{\mathbf{v}_i\}_{i=1}^H = \Psi_{\mathrm{CLS}}(\mathcal{I}),
\quad
\{\mathbf{v}_i\}_{i=1}^H = \Psi_{\mathrm{MLP}}(\mathcal{I})
\]
Noting, both schemes only adds a small number of parameters (negligible for $\Psi_{\mathrm{CLS}}$, $3\% \times H$ for $\Psi_{\mathrm{MLP}}$), so cost and speed for training and inference remain almost the same.

After obtaining $\{\mathbf{v}_i\}_{n=1}^H$, we match it with the text embedding sequence $\{\mathbf{t}_i\}_{i=1}^M$. Normally, we set $M=H$ to achieve part-by-part matching, while for larger $M$, we set $H<M$ and do grouped matching or free matching. We explain more details on asymmetric matching in supplementary materials. To optimize such one pipeline, we update to introduce multi-to-multi contrastive learning accordingly:
\begin{equation} 
\mathcal{L}^{\mathrm{T2I}}_{\text{M2M}} = - \sum_{i=1}^K \sum_{j=1}^M \log \frac{\exp(\langle \mathbf{v}_{i,j}, \mathbf{t}_{i,j} \rangle / \tau)}{\sum_{k=1}^K \exp(\langle \mathbf{v}_{k,j}, \mathbf{t}_{i,j} \rangle / \tau)}.
\end{equation}
The final optimization loss is $\mathcal{L}_{\text{M2M}} = (\mathcal{L}^{\mathrm{T2I}}_{\text{M2M}} + \mathcal{L}^{\mathrm{I2T}}_{\text{M2M}})/2$.

\vspace{0.1cm}
\noindent \textbf{Inference}. 
Unlike one-to-one CLIP, by one forward pass, we obtain $H$ normalized image embeddings $\{\mathbf{v}_i\}_{n=1}^H$. To aggregate them for downstream tasks, there are many strategies, {\em e.g.}, matching text embeddings with each image embeddings first, then taking max, norm max, or average (see Table~\ref{tab:ablation} for details). Normally, we simply use the average to strike the balance between efficiency and performance. Besides, an additional benefit is our M2M learning well disentangles image embeddings, comparing to O2M learning, which allows for flexible embedding customizations when reasoning for specific tasks (see Table~\ref{tab:deform}).

\section{Experiment} \label{sec:exp}
\subsection{Benchmark \& Evaluation}
\noindent \textbf{Training Datasets}. \underline{CC3M}~\cite{sharma2018conceptual} and \underline{CC12M}~\cite{changpinyo2021conceptual} are widely used vision+language pre-training, due to their large-scale size and cross-modal diversity. They collect massive web images, and alt-texts as initialization, then adopt a series of post-processing, {\em e.g.}, image filtering, text cleaning for final (image, text) version. Still, most images from these datasets are described by one coarse-grained text, in the form of monotonous short sentence. And sometimes, such text contains considerable noise, due to the captionor's bias.

\begin{table*}[t]
\centering
\footnotesize
\caption{\textbf{Complex Long-Text Retrieval.} On all datasets, our holistic CLIP, {\em i.e.}, $\Psi_{\mathrm{CLS}}$ and $\Psi_{\mathrm{MLP}}$, shows the gratifying superiority. Using the same data, our M2M paradigm surpasses recent O2M learning~\cite{fan2024improving,zheng2025dreamlip,liu2023mllms} across all metrics, showing the superiority.}
\vspace{-0.3cm}
\begin{tabular}{cC{0.9CM}|C{1.1CM}|C{0.5CM}C{0.6CM}C{0.5CM}|C{0.4CM}C{0.4CM}C{0.6CM}C{0.4CM}C{0.4CM}C{0.7CM}|C{0.4CM}C{0.4CM}C{0.6CM}C{0.4CM}C{0.4CM}C{0.7CM}}
\toprule
\multicolumn{2}{c|}{\multirow{3}{*}{Method}} & \multirow{3}{*}{\begin{tabular}[c]{@{}c@{}}Training\\ Text\end{tabular}} & \multicolumn{3}{c|}{\multirow{2}{*}{Caption}} & \multicolumn{6}{c|}{Fashion200-5K} & \multicolumn{6}{c}{Sharegpt4-5K} \\ \cline{7-18} 
\multicolumn{2}{c|}{} &  & \multicolumn{3}{c|}{} & \multicolumn{3}{c}{I2T} & \multicolumn{3}{c|}{T2I} & \multicolumn{3}{c}{I2T} & \multicolumn{3}{c}{T2I} \\ \cline{4-18} 
\multicolumn{2}{c|}{} &  & VLM & Prompt & Num & R@1 & R@5 & R@10 & R@1 & R@5 & R@10 & R@1 & R@5 & R@10 & R@1 & R@5 & R@10 \\ \hline \hline
\multicolumn{2}{c|}{CLIP~\cite{clip}} & CC3M & - & - & 1 & 2.2 & 6.2 & 9.5 & 2.6 & 7.4 & 11.8 & 11.0 & 25.9 & 37.1 & 10.2 & 26.0 & 36.4 \\ \hline
\multicolumn{2}{c|}{\multirow{3}{*}{O2M~\cite{fan2024improving,zheng2025dreamlip,liu2023mllms}}} & \multirow{3}{*}{\begin{tabular}[c]{@{}c@{}}CC3M\\+Caption\end{tabular}} & Uni & Uni & 2 & 5.5 & 14.1 & 21.3 & 5.5 & 15.8 & 23.3 & 39.6 & 65.8 & 76.2 & 38.3 & 65.0 & 75.1 \\
\multicolumn{2}{c|}{} &  & Uni & Multi & 5 & 8.8 & 21.8 & 30.6 & 9.5 & 22.4 & 30.8 & 48.4 & 75.6 & 84.2 & 49.2 & 75.7 & 84.3 \\
\multicolumn{2}{c|}{} &  & Multi & Uni & 5 & 8.4 & 21.4 & 30.0 & 7.7 & 21.3 & 29.3 & 48.6 & 73.8 & 82.7 & 49.2 & 75.4 & 83.6 \\ \hline 
\multicolumn{1}{c|}{\multirow{6}{*}{Ours}} & \multirow{3}{*}{$\Psi_{\mathrm{CLS}}$} & \multirow{3}{*}{\begin{tabular}[c]{@{}c@{}}CC3M\\+Caption\end{tabular}} & Uni & Uni & 2 & 5.5 & 15.2 & 21.7 & 6.1 & 16.5 & 23.8 & 41.3 & 66.7 & 77.0 & 39.9 & 66.3 & 76.2 \\
\multicolumn{1}{c|}{} &  &  & Uni & Multi & 5 & 9.0 & 23.0 & 32.0 & 9.5 & 24.4 & 32.8 & 51.2 & 76.8 & 85.4 & 52.4 & 77.5 & 85.5 \\
\multicolumn{1}{c|}{} &  &  & Multi & Uni & 5  & 9.2 & 22.2 & 30.4 & 9.0 & 22.2 & 31.0 & 49.4 & 75.1 & 83.3 & 50.6 & 76.9 & 84.4 \\ \cline{2-18} 
\multicolumn{1}{c|}{} & \multirow{3}{*}{$\Psi_{\mathrm{MLP}}$} & \multirow{3}{*}{\begin{tabular}[c]{@{}c@{}}CC3M\\+Caption\end{tabular}} & Uni & Uni & 2 & 5.6 & 16.3 & 23.6 & 6.2 & 17.3 & 24.6 & 41.4 & 67.6 & 77.3 & 39.6 & 66.6 & 76.5 \\
\multicolumn{1}{c|}{} &  &  & Uni & Multi & 5 & 9.3 & 23.6 & 31.4 & 9.6 & 22.6 & 31.4 & 53.0 & 77.7 & 85.9 & 53.0 & 78.5 & 86.2 \\
\multicolumn{1}{c|}{} &  &  & Multi & Uni & 5 & 9.2 & 22.4 & 31.2 & 8.9 & 22.1 & 31.1 & 50.5 & 75.8 & 83.6 & 51.4 & 77.2 & 85.6 \\ \hline \hline
\multicolumn{2}{c|}{CLIP~\cite{clip}} & CC12M & - & - & 1 & 8.6 & 22.4 & 31.6 & 8.5 & 21.9 & 31.4 & 33.9 & 61.3 & 71.3 & 32.9 & 60.1 & 71.4 \\  \hline 
\multicolumn{2}{c|}{\multirow{3}{*}{O2M~\cite{fan2024improving,zheng2025dreamlip,liu2023mllms}}} & \multirow{3}{*}{\begin{tabular}[c]{@{}c@{}}CC12M\\+Caption\end{tabular}} & Uni & Uni & 2 & 12.2 & 28.7 & 38.8 & 11.5 & 28.0 & 38.3 & 54.5 & 78.2 & 86.0 & 53.9 & 79.0 & 87.0 \\
\multicolumn{2}{c|}{} &  & Uni & Multi & 5 & 13.2 & 29.5 & 39.6 & 13.8 & 31.2 & 41.3 & 54.8 & 80.2 & 87.8 & 56.0 & 81.2 & 88.7 \\
\multicolumn{2}{c|}{} &  & Multi & Uni & 5 & 15.9 & 35.0 & 45.2 & 16.0 & 35.1 & 46.2 & 61.0 & 83.5 & 90.2 & 62.4 & 84.6 & 90.6 \\ \hline 
\multicolumn{1}{c|}{\multirow{6}{*}{Ours}} & \multirow{3}{*}{$\Psi_{\mathrm{CLS}}$} & \multirow{3}{*}{\begin{tabular}[c]{@{}c@{}}CC12M\\+Caption\end{tabular}} & Uni & Uni & 2 & 11.8 & 28.8 & 39.1 & 12.5 & 29.7 & 39.6 & 54.9 & 79.0 & 86.9 & 53.7 & 79.6 & 87.5 \\
\multicolumn{1}{c|}{} &  &  & Uni & Multi & 5 & 14.9 & 33.6 & 43.3 & 16.1 & 35.4 & 45.5 & 62.2 & 84.6 & 91.0 & 62.8 & 85.9 & 91.5 \\
\multicolumn{1}{c|}{} &  &  & Multi & Uni & 5 & 17.3 & 36.7 & 46.9 & 17.8 & 38.7 & 49.3 & 61.5 & 83.9 & 90.6 & 63.0 & 85.7 & 91.8 \\  \cline{2-18} 
\multicolumn{1}{c|}{} & \multirow{3}{*}{$\Psi_{\mathrm{MLP}}$} & \multirow{3}{*}{\begin{tabular}[c]{@{}c@{}}CC12M\\+Caption\end{tabular}} & Uni & Uni & 2 & 12.2 & 29.2 & 39.6 & 11.9 & 28.7 & 39.0 & 55.1 & 78.7 & 86.5 & 54.5 & 79.5 & 87.0 \\
\multicolumn{1}{c|}{} &  &  & Uni & Multi & 5 & 16.7 & 35.7 & 45.8 & 16.4 & 35.3 & 46.1 & 63.1 & 85.6 & 91.0 & 64.4 & 86.2 & 91.7 \\
\multicolumn{1}{c|}{} &  &  & Multi & Uni & 5 & 17.2 & 36.6 & 46.6 & 17.7 & 37.8 & 49.0 & 62.8 & 83.4 & 90.5 & 62.8 & 85.7 & 91.4 \\ \bottomrule
\end{tabular}
\label{tab:longtext retrieval}
\end{table*}

\vspace{0.1cm}
\noindent \textbf{Re-Caption}. We use powerful VLMs: InternVL2-8B~\cite{chen2024internvl}, Minigpt4-7B~\cite{zhu2023minigpt}, LLaVA1.5-13B~\cite{liu2024improved}, QwenVL2-7B~\cite{bai2023qwen} to generate diverse (image, multi-texts) pairs. When utilizing multi-VLMs, we set the unified prompt refer to~\cite{liu2023mllms}, ensuring the high quality of captioned texts. When adopting one VLM, we choose the classical InternVL2, and offer more details about the defined prompts and the statistics of captioned texts in supplementary materials.

\vspace{0.1cm}
\noindent \textbf{Testing Datasets \& Metrics}. For image classification, we employ ImageNet~\cite{deng2009imagenet}(-A/-R) validation set. For shot-text retrieval, MSCOCO~\cite{lin2014microsoft} and Flickr30~\cite{young2014image} are two typical benchmarks. For long-context retrieval, we randomly select 5K data from Fashion200K~\cite{han2017automatic} and Sharegpt4~\cite{chen2023sharegpt4v} to form Fashion200-5K and Sharegpt4-5K. To test the overall quality of the embedding sequence, we train ClipCap~\cite{mokady2021clipcap} for image captioning and LLaVA1.5-13B~\cite{liu2024improved} for all-round evaluations, by replacing the vision backbone with our image encoder. Note that, all above testing sets are standardized (without re-captioning) for the sake of fairnesss.

We follow conventions for evaluation: using the Top-1/5 accuracy for classification; B@4 and CIDEr for captioning; R@1/5/10 for image-text retrieval; and follow~\cite{duan2024vlmevalkit} to evaluate LLaVA on~\cite{liu2025mmbench,li2024seed,lu2022learn,schwenk2022okvqa,singh2019towards,Guan_2024_CVPR,yu2024mm,fu2023mme,mishraICDAR19}.

\subsection{Comparison with State-Of-The-Art}
Across many downstream tasks, we compare fully to vanilla CLIP~\cite{clip} and its variants~\cite{zheng2025dreamlip,fan2024improving,liu2023mllms} that use one-to-multi (O2M) contrastive learning to show our superiority.

\begin{table*}[t]
\caption{\textbf{Zero-shot Image Classification on CC12m.} Both diverse (image, multi-texts) data and M2M learning bring clear gains.}
\vspace{-0.3cm}
\centering
\footnotesize
\begin{tabular}{cc|c|ccc|cccccc}
\toprule
\multicolumn{2}{c|}{\multirow{2}{*}{Method}} & \multirow{2}{*}{\begin{tabular}[c]{@{}c@{}}Training\\ Text\end{tabular}} & \multicolumn{3}{c|}{Caption} & \multicolumn{2}{c}{ImageNet} & \multicolumn{2}{c}{ImageNet-A} & \multicolumn{2}{c}{ImageNet-R} \\ \cline{4-12} 
\multicolumn{2}{c|}{} &  & VLM & Prompt & Num & Top1 & Top5 & Top1 & Top5 & Top1 & Top5 \\ \hline \hline
\multicolumn{2}{c|}{CLIP~\cite{clip}} & CC12M & - & - & 1 & 39.0 & 66.9 & 10.1 & 33.2 & 49.2 & 74.1 \\ \hline
\multicolumn{2}{c|}{\multirow{3}{*}{O2M~\cite{fan2024improving,zheng2025dreamlip,liu2023mllms}}} & \multirow{3}{*}{\begin{tabular}[c]{@{}c@{}}CC12M\\ +Caption\end{tabular}} & Uni & Uni & 2 & 43.1 & 71.4 & 12.1 & 36.8 & 54.7 & 78.8 \\
\multicolumn{2}{c|}{} &  & Uni & Multi & 5 & 45.8 & 74.3 & 17.6 & 46.4 & 59.6 & 81.4 \\ 
\multicolumn{2}{c|}{} &  & Multi & Uni & 5 & 46.5 & 75.0 & 16.0 & 44.9 & 60.7 & 83.5 \\ \hline
\multicolumn{1}{c|}{\multirow{6}{*}{Ours}} & \multirow{3}{*}{$\Psi^{\mathrm{CLS}}$} & \multirow{3}{*}{\begin{tabular}[c]{@{}c@{}}CC12M\\ +Caption\end{tabular}} & Uni & Uni & 2 & 43.5 & 71.6 & 12.0 & 36.5 & 54.9 & 79.2 \\
\multicolumn{1}{c|}{} &  &  & Uni & Multi & 5 & 48.6 & 76.6 & 18.4 & 47.5 & 62.1 & 84.6 \\
\multicolumn{1}{c|}{} &  &  & Multi & Uni & 5 & 46.8 & 75.2 & 16.8 & 44.8 & 65.0 & 86.0 \\ \cline{2-12} 
\multicolumn{1}{c|}{} & \multirow{3}{*}{$\Psi^{\mathrm{MLP}}$} & \multirow{3}{*}{\begin{tabular}[c]{@{}c@{}}CC12M\\ +Caption\end{tabular}} & Uni & Uni & 2 & 43.0 & 71.0 & 11.2 & 35.2 & 54.7 & 79.0 \\
\multicolumn{1}{c|}{} &  &  & Uni & Multi & 5 & 48.1 & 76.6 & 19.0 & 48.1 & 61.6 & 84.0 \\ 
\multicolumn{1}{c|}{} &  &  & Multi & Uni & 5 & 46.6 & 75.3 & 16.3 & 44.7 & 63.9 & 85.5 \\ \bottomrule
\end{tabular}
\label{tab:Zeroshot-ImageClassification}
\end{table*}

\begin{table}[t]
\centering
\footnotesize
\caption{\textbf{Image-to-Text Captioning}. We train ClipCap~\cite{mokady2021clipcap} using our holistic CLIP to show the power of M2M contrastive learning.}
\vspace{-0.3cm}
\begin{tabular}{C{4CM}C{1.3CM}|C{1.3CM}|C{1CM}C{1CM}}
\toprule
\multicolumn{2}{c|}{\multirow{2}{*}{Method}} & \multirow{2}{*}{\begin{tabular}[c]{@{}c@{}}Training\\ Text\end{tabular}} & \multicolumn{2}{c}{COCO Caption} \\ \cline{4-5} 
\multicolumn{2}{c|}{} &  & B@4 & CIDEr \\ \hline \hline
\multicolumn{2}{c|}{O2M~\cite{fan2024improving,zheng2025dreamlip,liu2023mllms}} & \multirow{3}{*}{CC3M} & 26.5 & 74.7 \\ \cline{1-2} \cline{4-5} 
\multicolumn{1}{c|}{\multirow{2}{*}{Ours}} & $\Psi_{\mathrm{CLS}}$ &  & 27.2 & 76.5 \\
\multicolumn{1}{c|}{} & $\Psi_{\mathrm{MLP}}$ &  & \textbf{27.6} & \textbf{76.7} \\ \hline \hline
\multicolumn{2}{c|}{O2M~\cite{fan2024improving,zheng2025dreamlip,liu2023mllms}} & \multirow{3}{*}{CC12M} & 27.0 & 76.8 \\ \cline{1-2} \cline{4-5} 
\multicolumn{1}{c|}{\multirow{2}{*}{Ours}} & $\Psi_{\mathrm{CLS}}$ &  & 28.6 & 81.6 \\
\multicolumn{1}{c|}{} & $\Psi_{\mathrm{MLP}}$ &  & \textbf{28.8} & \textbf{82.1} \\ \bottomrule
\end{tabular}
\label{tab:captioning}
\end{table}

\begin{table*}[t]
\centering
\footnotesize
\caption{\textbf{All-Round Embedding Quality on Whole Visual Token Sequence.} We train LLaVA1.5-13B by replacing its vision backbone by our image encoder of pre-trained holistic CLIP $\Psi_{\mathrm{CLS}}$, and the good results show better embeddings along the whole visual sequence.}
\vspace{-0.3cm}
\begin{tabular}{c|c|c|C{1cm}C{0.6cm}cC{0.8cm}ccccC{0.8cm}c}
\hline
Method & \begin{tabular}[c]{@{}c@{}}Training\\ Text\end{tabular} & \begin{tabular}[c]{@{}c@{}}Caption\\ Prompt\end{tabular} & \begin{tabular}[c]{@{}c@{}}MMBench\\ EN~\cite{liu2025mmbench}\end{tabular} & \begin{tabular}[c]{@{}c@{}}CC\\ Bench\end{tabular} & MME & \begin{tabular}[c]{@{}c@{}}A-\\ OKVQA\end{tabular} & \begin{tabular}[c]{@{}c@{}}Hallusion\\ Bench\end{tabular} & MMVeT & \begin{tabular}[c]{@{}c@{}}OCR\\ VQA\end{tabular} & \begin{tabular}[c]{@{}c@{}}SEED\\ BenchIMG\end{tabular} & \begin{tabular}[c]{@{}c@{}}Science\\ QA\end{tabular} & \begin{tabular}[c]{@{}c@{}}Text\\ VQA\end{tabular} \\ \hline  \hline
O2M & \multirow{2}{*}{\begin{tabular}[c]{@{}c@{}}CC3M\\ +Caption\end{tabular}} & \multirow{2}{*}{Multi} & 41.2 & 2.7 & 842.8 & 60.2 & 29.2 & 15.7 & \textbf{26.4} & 44.4 & 63.9 & 2.2 \\
Ours &  &  & \textbf{41.7} & \textbf{2.9} & \textbf{944.6} & \textbf{63.2} & \textbf{32.2} & \textbf{16.4} & 26.3 & \textbf{47.9} & \textbf{64.1} & \textbf{3.0} \\ \hline
O2M & \multirow{4}{*}{\begin{tabular}[c]{@{}c@{}}CC12M\\ +Caption\end{tabular}} & \multirow{2}{*}{Uni}  & 42.1 & 3.3 & 909.5 & 62.6 & 34.5 & 17.4 & 27.6 & 48.0 & 63.9 & 3.7 \\
Ours &  & & 42.3 & 3.0 & 930.2 & 61.1 & 34.3 & 18.5 & 27.5 & 45.4 & 63.9 & 3.8 \\ \cline{1-1} \cline{3-13} 
O2M &  & \multirow{2}{*}{Multi} & 42.6 & 3.3 & 966.7 & 61.2 & 34.3 & 17.9 & 27.1 & 45.5 & 64.6 & 4.4 \\
Ours &  & & \textbf{50.3} & \textbf{5.9} & \textbf{997.5} & \textbf{64.5} & \textbf{38.9} & \textbf{19.7} & \textbf{27.8} & \textbf{51.2} & \textbf{65.4} & \textbf{6.6}  \\ \hline
\end{tabular}
\label{tab:llava}
\end{table*}

\begin{table}[t]
\centering
\footnotesize
\caption{\textbf{Ablation Study for the Embedding Fusion.} The AVG-based one surpasses others in both performance and efficiency.}
\vspace{-0.3cm}
\begin{tabular}{c|cc|cc|c}
\toprule
\multirow{2}{*}{\begin{tabular}[c]{@{}c@{}}Fusion\\ Strategy\end{tabular}} & \multicolumn{2}{c|}{COCO} & \multicolumn{2}{c|}{Flickr30} & \multicolumn{1}{c}{ImageNet-R} \\
 & I2T & T2I & I2T & T2I & Top-1 \\
\hline \hline
Max & 36.3 & 35.8 & 74.0 & 73.8 & 60.5 \\
Norm-Max & 36.1 & 35.2 & 74.5 & 74.7 & 60.9 \\
Average & \textbf{38.2} & \textbf{38.2} & \textbf{76.6} & \textbf{76.5} & \textbf{61.6} \\
\bottomrule
\end{tabular}
\label{tab:ablation}
\end{table}

\begin{table}[t]
\centering
\footnotesize
\caption{\textbf{Embedding Customization}. On CC12M, we decompose embeddings and select parts to combine for specific tasks . `Main-Obj' focuses on main objects, `Long-Text' matches rich details.}
\vspace{-0.3cm}
\begin{tabular}{c|cc|cc|c|c}
\toprule
\multirow{2}{*}{\begin{tabular}[c]{@{}c@{}}Embedding\\ Focus\end{tabular}} & \multicolumn{2}{c|}{COCO} & \multicolumn{2}{c|}{Sharegpt4-5k} & \multicolumn{1}{c|}{ImgNet} & \multicolumn{1}{c}{ImgNet-R} \\
 & I2T & T2I & I2T & T2I & Top-1 & Top-1 \\
\hline \hline
All & 38.2 & \textbf{38.2} & 63.1 & 64.4 & 48.1 & 61.6 \\
Main Obj & 37.7 & 36.5 & 63.1 & 63.4 & \textbf{49.4} & \textbf{63.0} \\
Long Text & \textbf{38.5} & 38.0 & \textbf{63.5} & \textbf{64.7} & 48.9 & 62.8 \\
\bottomrule
\end{tabular}
\label{tab:deform}
\end{table}

\begin{table}[t]
\centering
\footnotesize
\caption{\textbf{Ablation Study for the Text Number on CC3M.} Since visual embeddings are decomposable, we can tear down to fit for specific tasks. The more captions used, the better the performance.}
\vspace{-0.3cm}
\begin{tabular}{C{1.1cm}|C{1.1cm}|C{0.5cm}C{0.5cm}|C{0.5cm}C{0.5cm}|C{1cm}}
\toprule
\multirow{2}{*}{\begin{tabular}[c]{@{}c@{}}Text\\ Number\end{tabular}} & 
\multirow{2}{*}{\begin{tabular}[c]{@{}c@{}}Text\\ Similarity\end{tabular}} & \multicolumn{2}{c|}{COCO} & \multicolumn{2}{c|}{Flickr30} & \multicolumn{1}{c}{ImageNet} \\
&  & I2T & T2I & I2T & T2I & Top-1 \\
\hline \hline
2  & 0.56 & 20.1 & 20.6 & 50.9 & 51.2 & 29.1 \\
3  & 0.52 & 25.3 & 25.4 & 58.8 & 58.3 & 33.7 \\
4  & 0.50 & 28.9 & 29.0 & 61.7 & 63.2 & 38.9 \\
5 & 0.48 & 28.2 & 27.4 & 63.7 & 63.9 & 38.8 \\
use 4 of 5 & 0.48  & 28.9 & 29.2 & 64.3 & 63.9 & 40.0 \\
\bottomrule
\end{tabular}
\label{tab:ablation1}
\end{table}

\vspace{0.1cm}
\noindent \textbf{Elementary Short-Text Retrieval} is evaluated in Table~\ref{tab:shorttext retrieval} on MSCOCO and Flickr30. We ablate both multi-to-multi (M2M) learning paradigm and (image, multi-texts) dataset, while one-to-multi (O2M) paradigm is used for comparison. Comparisons between M2M and O2M reveal consistent improvements across all multi-texts datasets, with gains of 1.5-3\% on multi-VLMs data and 0.3-1.5\% on multi-prompts data. Notably, the multi-prompts data improve more prominent than multi-VLMs data, achieving gains of 3-6\% across datasets through either $\Psi_{\mathrm{CLS}}$ or $\Psi_{\mathrm{MLP}}$. Such a remarkable gain underscores the effectiveness of our M2M paradigm for refining the embedding space, ensuring that it flexibly decomposes then aligns colorful vision to diverse texts. By enabling holistic alignments, M2M fundamentally mitigates mis-clustering and demonstrates robust resistance to potential text noise, providing the cohesive embedding space.

To explicitly enforce text diversity from VLMs, we employ a variety of prompts defined by multi-view and multi-level spirits. Compared to training on uni-view textual captions generated by InternVL2-8B, our solution yields significant improvements, showing 8-10\% gains on Flickr30 and 5-7\% boosts on MSCOCO. Notably, our multi-prompt results are on par with those trained on multi-VLMs datasets, which rely on more auxiliary knowledge and are tested on simplified datasets (comprised of general, rather than multi-view texts). These results suggest that our (image, multi-texts) datasets, combined with M2M contrastive learning, foster one more robust embedding space.

\vspace{0.1cm}
\noindent \textbf{Complex Long-Text Retrieval} is implemented in Table~\ref{tab:longtext retrieval} on the Fashion200-5K and Sharegpt4v-5K datasets, to evaluate the VLPs' efficacy in practical scenarios. In terms of pre-training optimization paradigm, the results illustrated that, M2M consistently outperforms O2M on (image, multi-texts) datasets, with gains of 4-7\% on Sharegpt4-5K and 1-4\% on Fashion200-5K. And noting that, Fashion200-5K is particularly complex, with $R@1$ scores only in the range of 9-15\%, making these improvements quite substantial.

When comparing models trained with (image, uni-text) versus (image, multi-texts) datasets, we observe marked advancements, with an 8\% gain on Sharegpt4-5K and a 4\% gain on Fashion200-5K. Besides that, to obtain high-quality (image, multi-texts) pairs, explicit prompt differences surpass implicit VLMs groups, underscoring the value of holistic data construction for enhanced performance.

\vspace{0.1cm}
\noindent \textbf{Zero-Shot Image Classification}. For Table~\ref{tab:Zeroshot-ImageClassification}, we present evaluations on several widely-recognized benchmarks for classification. While M2M contrastive learning consistently yields substantial improvements, it is particularly noteworthy that models trained on the multi-prompts and uni-VLM setting outperform the multi-VLMs and uni-prompt setting. This further substantiates the critical role of comprehensive textual alignment in advancing model performance.

\vspace{0.1cm}
\noindent \textbf{Image-to-Text Captioning}. Table~\ref{tab:captioning} assesses the effectiveness of our holistic CLIP, for dense tasks such as captioning. We adapt ClipCap~\cite{mokady2021clipcap} by substituting its vision prefix through our pre-trained visual encoder. On both CC3M and CC12M datasets, consistent performance improvements are revealed by training with (image, multi-texts) pairs, again demonstrating that M2M paradigm's outstanding capability in effectively capturing and integrating semantics.

\vspace{0.1cm}
\noindent \textbf{All-Round Abilities}. To further demonstrate our superiority for the whole visual sequence, we leverage LLaVA1.5-13B as the VLM and replace its visual backbone with pre-trained visual encoder of holistic CLIP $\Psi_{\mathrm{CLS}}$. Table~\ref{tab:llava} evaluates on a diverse set of benchmarks~\cite{duan2024vlmevalkit}. Both M2M learning and (image, multi-texts) data produce significant advantages. Across nearly all the benchmarks, M2M consistently outperforms O2M, indicating that M2M not only captures powerful embedding summaries, but also enhances quality for the whole visual token sequence. Moreover, multi-view, multi-level texts markedly strengthen model’s abilities, further validating the efficacy of our idea.

\subsection{Ablation Study}
To form our holistic CLIP pipeline, M2M contrastive learning and (image, multi-texts) paired data are the keys. Next, we dissect their effectiveness.

\vspace{0.1cm}
\noindent \textbf{Efficacy of Holistic Image-Text Data}. We present models trained with (image, uni-text) pairs and (image, multi-texts) pairs in Table~\ref{tab:shorttext retrieval},~\ref{tab:longtext retrieval},~\ref{tab:Zeroshot-ImageClassification}, and~\ref{tab:llava}. In terms of performance, using prompts of diversified definitions to steer one VLM (InternVL2) reaches comparable results with using one prompt to drive substantially larger VLM pools, and both show significant gains. On object-oriented image classification and long-context retrieval, the former performs even better.

\vspace{0.1cm}
\noindent \textbf{Efficacy of Multi-to-Multi Contrastive Learning}. Across various benchmarks using the same dataset, M2M consistently outperforms O2M, with especially notable gains under (image, multi-texts) settings. With the increase in text diversity (caption number from 2 to 5), the advantages of M2M are gradually becoming apparent, which is expected, given that M2M is designed for multi-to-multi matching. Table~\ref{tab:llava} further illustrates that M2M also strengthens embedding learning across the whole visual sequence.

Moreover, $\Psi_{\mathrm{CLS}}$ and $\Psi_{\mathrm{MLP}}$ bring similar performance across all benchmarks, suggesting M2M contrastive learning itself is the principal driver behind these improvements.

\vspace{0.1cm}
\noindent \textbf{Fusion During Inference}. When reasoning, we integrate the embedding set learned by the holistic CLIP to perform image-text matching tasks. In Table~\ref{tab:ablation}, three common fusion strategies are assessed, and the AVG-based fusion outperforms the others, in terms of performance. For the actual efficiency, the AVG-based one also wins: when calculating cosine similarity, we can first average on image embeddings, then take dot product with text embeddings, thus avoiding additional computational cost.

\vspace{0.1cm}
\noindent \textbf{Flexible Embedding Disentanglement}. In above tables, we average all image embeddings for various tasks. Actually, as image embeddings can be disentangled, we can even customize them during inference, {\em e.g.}, choosing partial embeddings that is suitable for one specific task, and then averaging. Table~\ref{tab:ablation1} explores three simple customizations, by employing image embeddings that focus on the main objects, rich details and all information. More complex customizations could be explored in the future. The results show that, utilizing different feature combinations can further improve performance, also providing good interpretability.

Moreover, this disentanglement inherently enhances robustness to noise, as it could easily distinguish noisy textual branches, and we manually disable them for inference, thereby minimizing adverse effects. Notably, ``noise'' is one relative concept, depending on specific tasks. Anyway, our method achieves flexibly denoising without the need for retraining or fine-tuning, further strengthening the robustness.

\vspace{0.1cm}
\noindent \textbf{Efficacy of Text Number}. Table~\ref{tab:ablation1} carries out experiments to further dissect the efficacy of text number. As the model is exposed to an increasing number of descriptive texts with differences, its performance continues to improve. By extending texts from four to five adding background descriptions, we strengthen the model's semantic representation. However, this expansion may impact performance on several downstream tasks. To address this, we decompose image embeddings and exclude background branch for inference, yielding optimal performance across tasks.

Moreover, as text number adds from 2 to 5, the average similarity between multiple texts for each image is continuously decreasing, showing the increases in text diversity.

\subsection{Visualization of Holistic Alignment}
\begin{figure}[t]
\begin{center}
\includegraphics[width=0.49\textwidth] {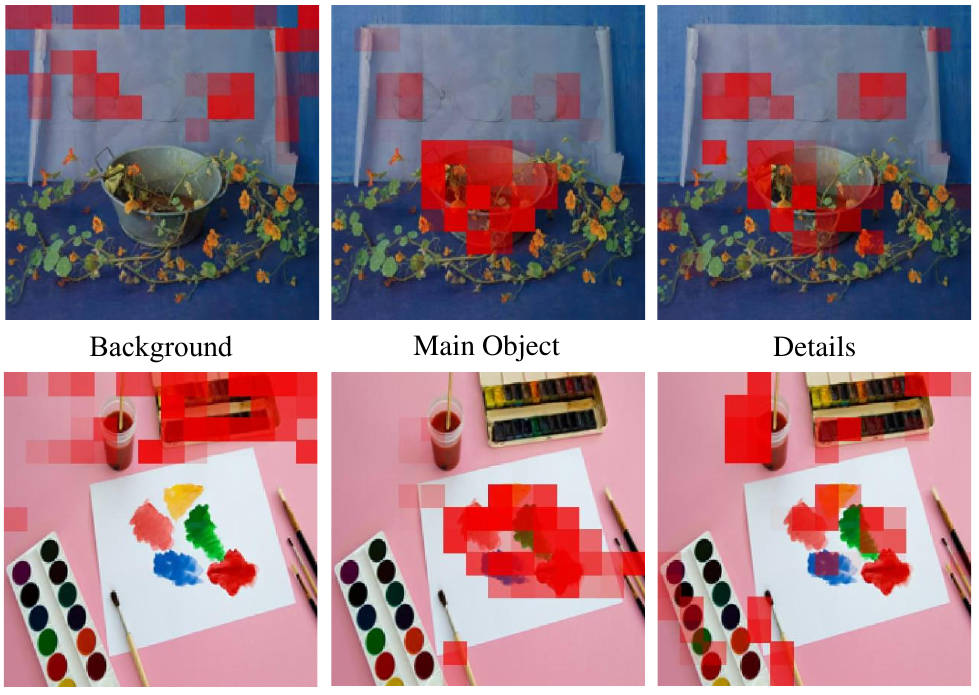}
\end{center}
\vspace{-0.6cm}
\caption{\textbf{Attention Visualization of Holistic CLIP's Vision.} Vision is naturally decomposed by aligning with various texts.}
\vspace{-0.3cm}
\label{fig:visualize}
\end{figure}

To evaluate the quality for learned visual embeddings, we highlight the top 20\% of visual patches with the highest attention scores in red, and darker colors indicating higher scores. As illustrated in Figure~\ref{fig:visualize}, distinct attention patterns emerge for class tokens guided by specific text types, and diverse visions of multi-view and multi-level are captured. Such alignment between visual attention and text perspective underscores the efficacy of our M2M constrastive learning; on the other hand, it is also shown that the performance is significantly affected by data diversity, emphasizing the importance of holistic (image, multi-texts) pairs.

\section{Conclusion}   \label{sec:conclusion}
This paper presents holistic CLIP to advance beyond myopic paradigms. By (image, multi-texts) data and multi-to-multi contrastive learning, we effectively improve the versatility and comprehensiveness of cross-modal embeddings. With extensive experiments and massive ablations, our idea shows superior performance in image-text retrieval, classification, many dense visual tasks, achieving notable gains over OpenAI's CLIP. This paper highlights the critical role of all-round aligned data and optimization for CLIP, inspiring the further prosperity of vision-language models.

{
\small
\bibliographystyle{ieeenat_fullname}
\bibliography{main}
}

\clearpage
\appendix
In supplementary materials, we first provide more details about our multi-to-multi contrastive learning; then offer the details to create (image, multi-texts) dataset by VLMs' captioning; finally demonstrate more visualizations.

\section{Implementation Details}
\subsection{Multi-to-Multi Contrastive Learning}
The image is much more fine-grained and informative than texts. To fully align the image semantics with text descriptions from various perspectives, we propose one novel contrastive learning paradigm called multi-to-multi contrastive learning (M2M). The objective of M2M is to let the model learn multi-perspective image features under the guidance of typical texts. M2M prevents the model from aligning one summary image feature with multiple text features with different semantic meanings.

\begin{figure*}[t]
\begin{center}
\includegraphics[width=0.98\textwidth] {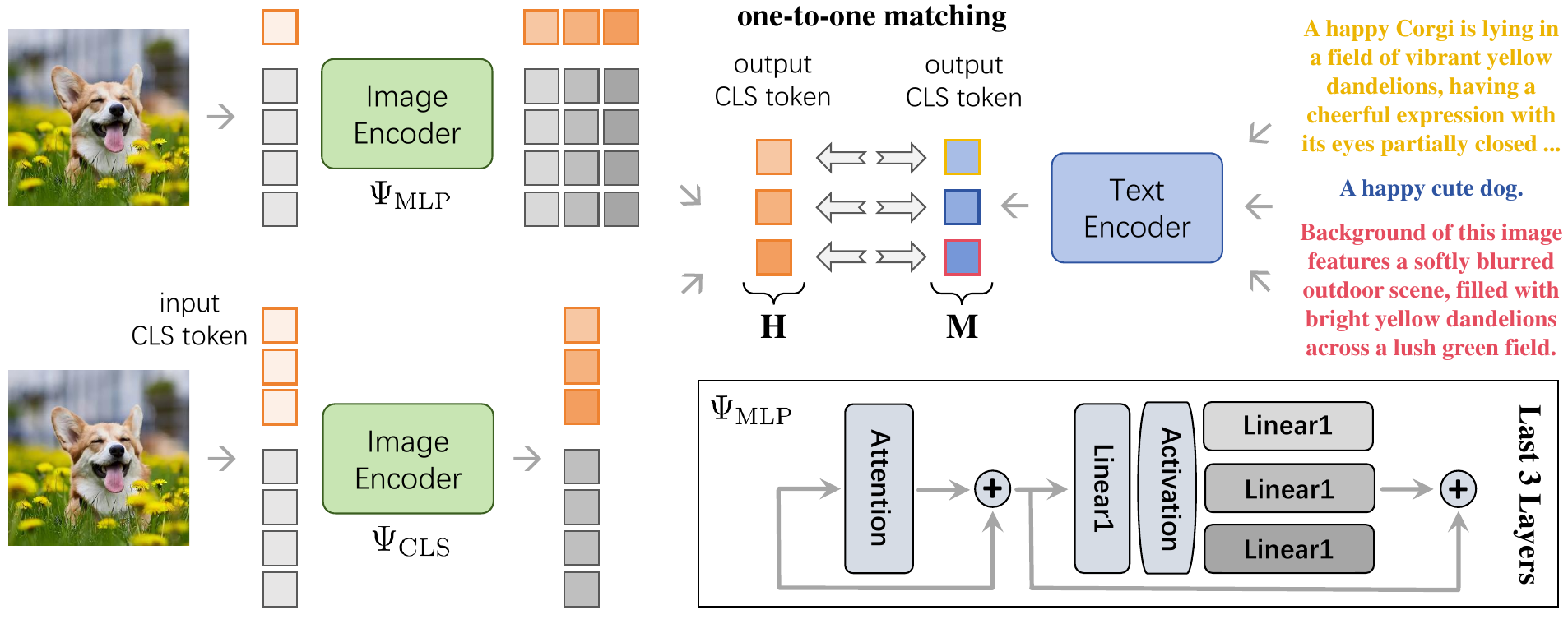}
\end{center}
\vspace{-0.5cm}
\caption{\textbf{Architecture Overview of Holistic CLIP.} To generate $H$ image features, we leverage two different structures: $\Psi_{\mathrm{CLS}}$ and $\Psi_{\mathrm{MLP}}$. Then we match $H$ image features with $M$ text features. Normally $H=M$ and we apply one-to-one matching.}
\label{fig:holistic}
\end{figure*}

\subsubsection{Holistic CLIP Model} 
To automatically output multi-branch image features in one forward pass, we design two types of architectures: class-token based $\Psi_{\mathrm{CLS}}$ and mlp-layer based $\Psi_{\mathrm{MLP}}$.

\vspace{0.1cm}
\noindent \textbf{$\mathbf{\Psi}_{\mathbf{\mathrm{CLS}}}$ Architecture}. For most VLMs, the image encoder is transformer-based architecture, thus we always a sequence of visual embeddings with one class token to summarize the overall semantics. However, in M2M pipeline, a single class token is not sufficient to cover representations from various perspectives (only one attention map for one class token). Thus, we initialize multiple class tokens, each focusing on specific aspects of an image guided by the assigned text descriptions. In this way, we obtain multiple image features with almost no extra efforts.

\vspace{0.1cm}
\noindent \textbf{$\mathbf{\Psi}_{\mathbf{\mathrm{MLP}}}$ Architecture}.
Although $\Psi_{\mathrm{CLS}}$ is already an effective way to extract features from different views, all the class tokens share the visual token sequence, which may not be always an ideal solution. Inspired by the MoE intuition~\cite{du2022glam}, we also introduce one MLP-based architecture. Specifically, to output $H$ image features, we expand the last 3 MLP layers to $H$ parallel parts. For efficiency, we only expand the second linear FFN. In Figure~\ref{fig:holistic}, we show in details the architecture of both $\Psi_{\mathrm{CLS}}$ and  $\Psi_{\mathrm{MLP}}$.

\subsubsection{Multi-to-Multi Matching}
Here, after obtaining the $H$ image features from $\{\mathbf{v}_i\}_{i=1}^H = \Psi_{\mathrm{CLS}}(\mathcal{I})$ or $\{\mathbf{v}_i\}_{i=1}^H = \Psi_{\mathrm{MLP}}(\mathcal{I})$ and $M$ text features from the corresponding multi-perspective texts $\{\mathbf{t}_i\}_{i=1}^M = \Phi^{\mathrm{txt}}(\{\mathcal{T}_i\}_{i=1}^M)$, we match them to jointly learn the alignment. Normally, we set $H=M$ so that we assign each image head a specific type of text to align (object-oriented, background, long caption...). If $M$ is relatively large, {\em e.g.}, if we prompt 4 different VLMs with 4 different prompts, thus to get 16 distinct texts. In this case we first group the texts into $H$ sets based on their similarities~\cite{reimers2019sentence}. If some texts are mixed up, we use free match: match the image head with maximum cosine similarity. After matching texts with their corresponding image features, we can train the model by M2M contrastive learning described in the main paper.

\subsubsection{Training Details}
We train ViT-B-16 model on 8 A100-80G GPUs with batch size 256 for 100 epochs on CC12M~\cite{changpinyo2021conceptual} and batch size 64 for 200 epochs on CC3M~\cite{changpinyo2021conceptual}, with a learning rate of $5e^{-4}$. As we tested, the above training setup achieves the best result on Conceptual datasets, though the training converges slower than larger batch size setup.

\subsubsection{Inference Details}
After generating multiple image features, we first normalize them, then take the average to produce one compositional image representation for retrieval-based downstream tasks. To train LLaVA1.5 for visually dense tasks, we replace the visual encoder by our trained model and use the whole output visual sequence as input for projection layer and LLM.

\begin{table*}[h]
\centering
\small
\setlength{\arrayrulewidth}{0.5pt} 
\renewcommand{\arraystretch}{1.5} 
\caption{\textbf{Multi-Prompts Used for Constructing (Image, Multi-Texts) Data.}}
\vspace{-0.3cm}
\begin{tabular}{c|c}
\hline
\textbf{Type} & \textbf{Prompt} \\
\hline
Details & $<$image$>$ Describe the image in detail. \\
Nouns & Briefly describe the image with few noun words separated by ``,''. \\
Main Object & Describe only the one main object in the image, do not say anything about the other objects or background. \\
Background & Describe only the background of this image, do not say anything about the foreground objects. \\
Style & Describe the style or your feelings about this image, do not say anything about the objects in the image. \\
\hline
\end{tabular}
\label{tab:prompt5}
\end{table*}

\section{(Image, Multi-Texts) Dataset}
\subsection{Prompts \& Examples}
\begin{figure*}[t]
\begin{center}
\includegraphics[width=0.95\textwidth] {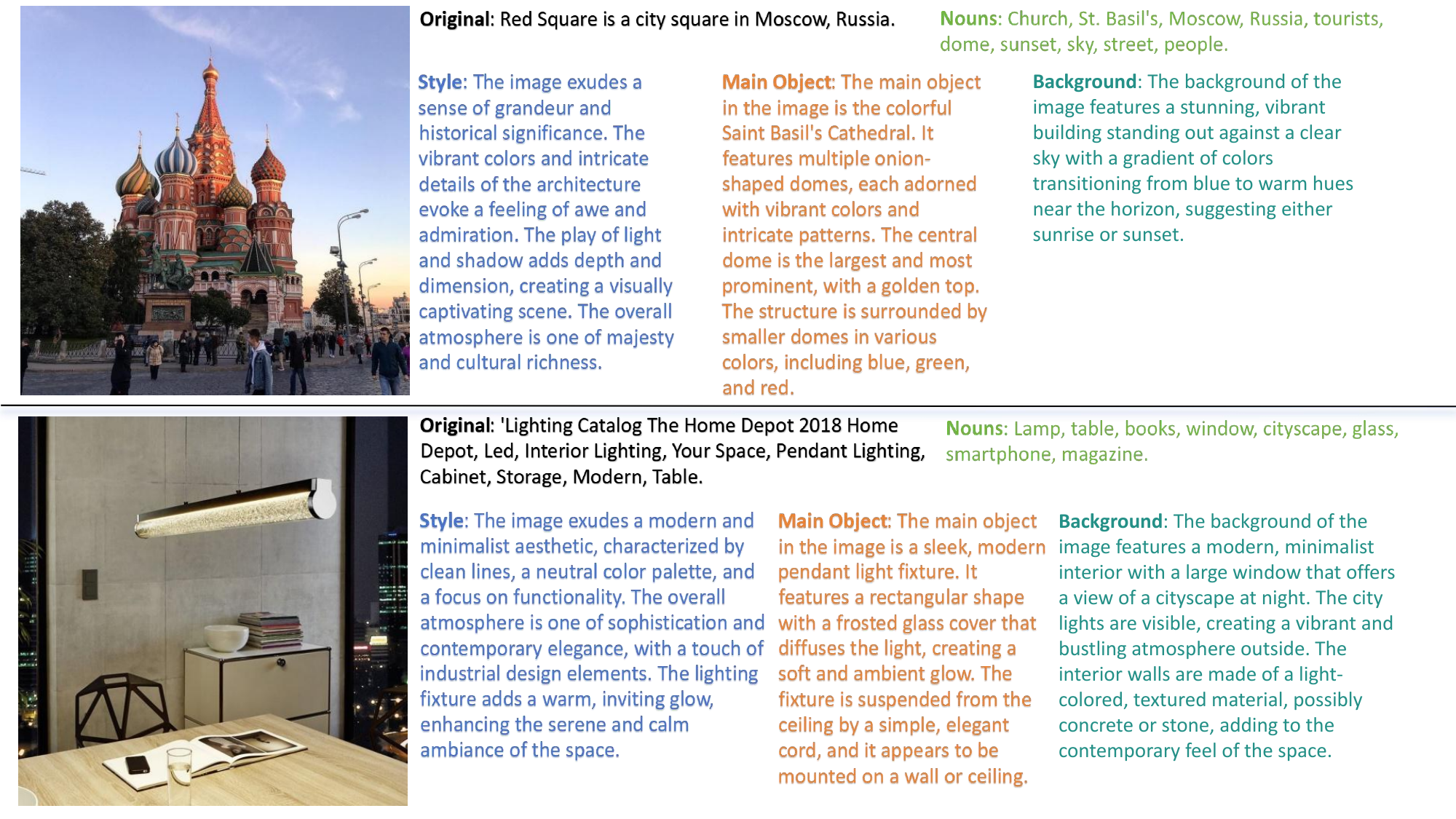}
\end{center}
\vspace{-0.6cm}
\caption{\textbf{Examples of (Image, Multi-Texts) Data from Multi-Prompts.}}
\label{fig:prompt1}
\end{figure*}

\begin{figure*}[t]
\begin{center}
\includegraphics[width=0.95\textwidth] {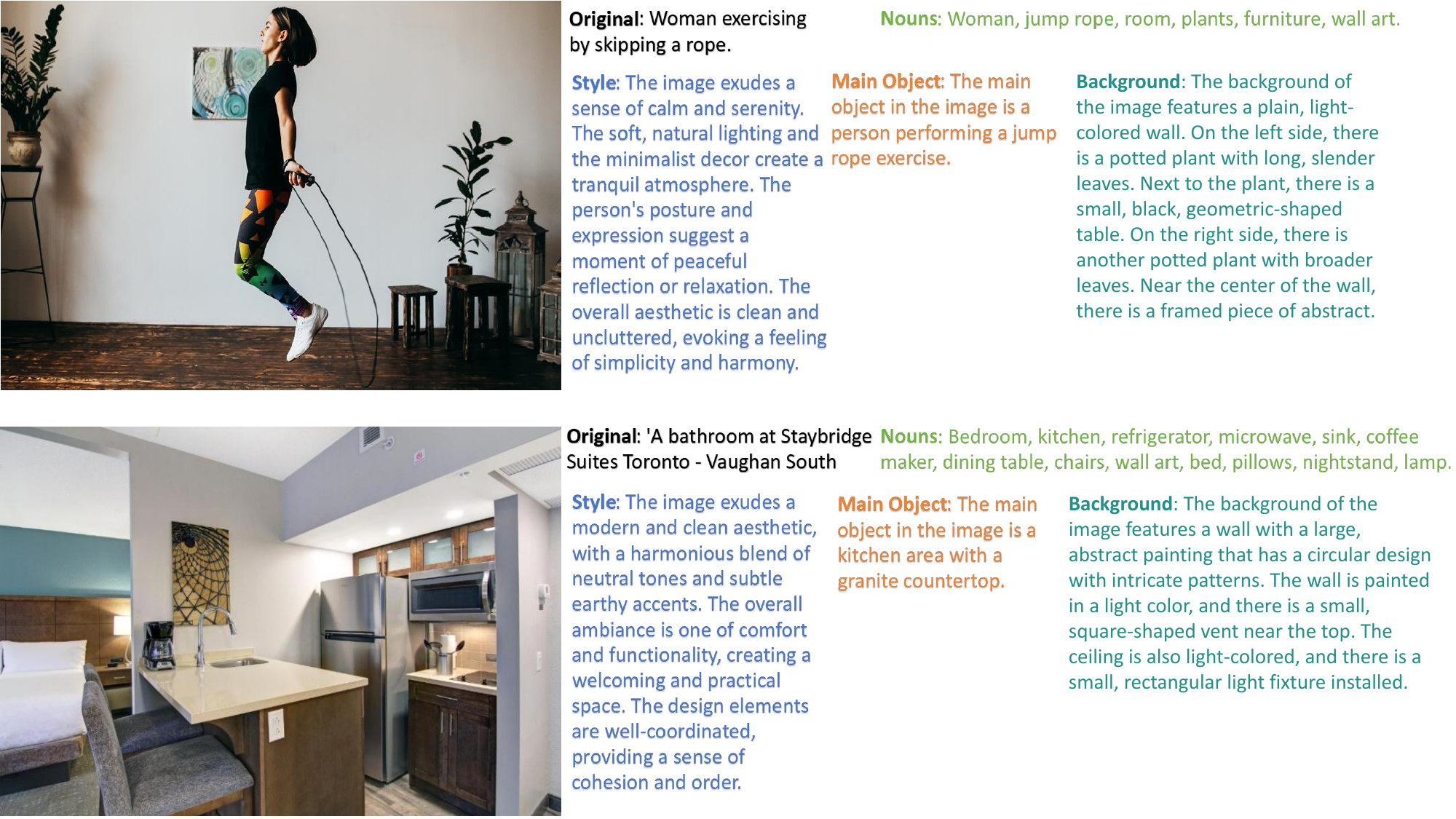}
\end{center}
\vspace{-0.6cm}
\caption{\textbf{Examples of (Image, Multi-Texts) Data from Multi-Prompts.}}
\label{fig:prompt2}
\end{figure*}

In the main paper, we describe the prompt design principle for multi-view/granularity captions: \underline{\textit{Focus Guide}} indicates whether attention should be laid on foreground objects or general background substances. \underline{\textit{Physical or Sensory}} refers to describe solid nouns or feeling styles. \underline{\textit{Gaze or Glance}} define captions are dense long-details or compact short-overview. \underline{\textit{Complex Reasoning}} distinguishes relationship or sequence for entities. To generate text descriptions from the above aspects, we design 5 different prompts, which are listed in Table~\ref{tab:prompt5}. We prompt InternVL2-8B to produce captions from various perspectives. In Figure~\ref{fig:prompt1} and~~\ref{fig:prompt2}, we offer some examples from CC12M~\cite{changpinyo2021conceptual} datasets. We can see that these texts describe the image from various perspectives, with different semantic meanings. For common one-to-multi contrastive learning, we align one summary image feature with all the text features, which inevitably pushes texts with different semantic meanings close to each other in the embedding space, making the alignment chaotic and confusing. However, our multi-to-multi contrastive learning fundamentally prevents such chaos, by outputting multiple image representations and learn alignment independently.

\begin{figure*}[t]
\begin{center}
\includegraphics[width=0.95\textwidth] {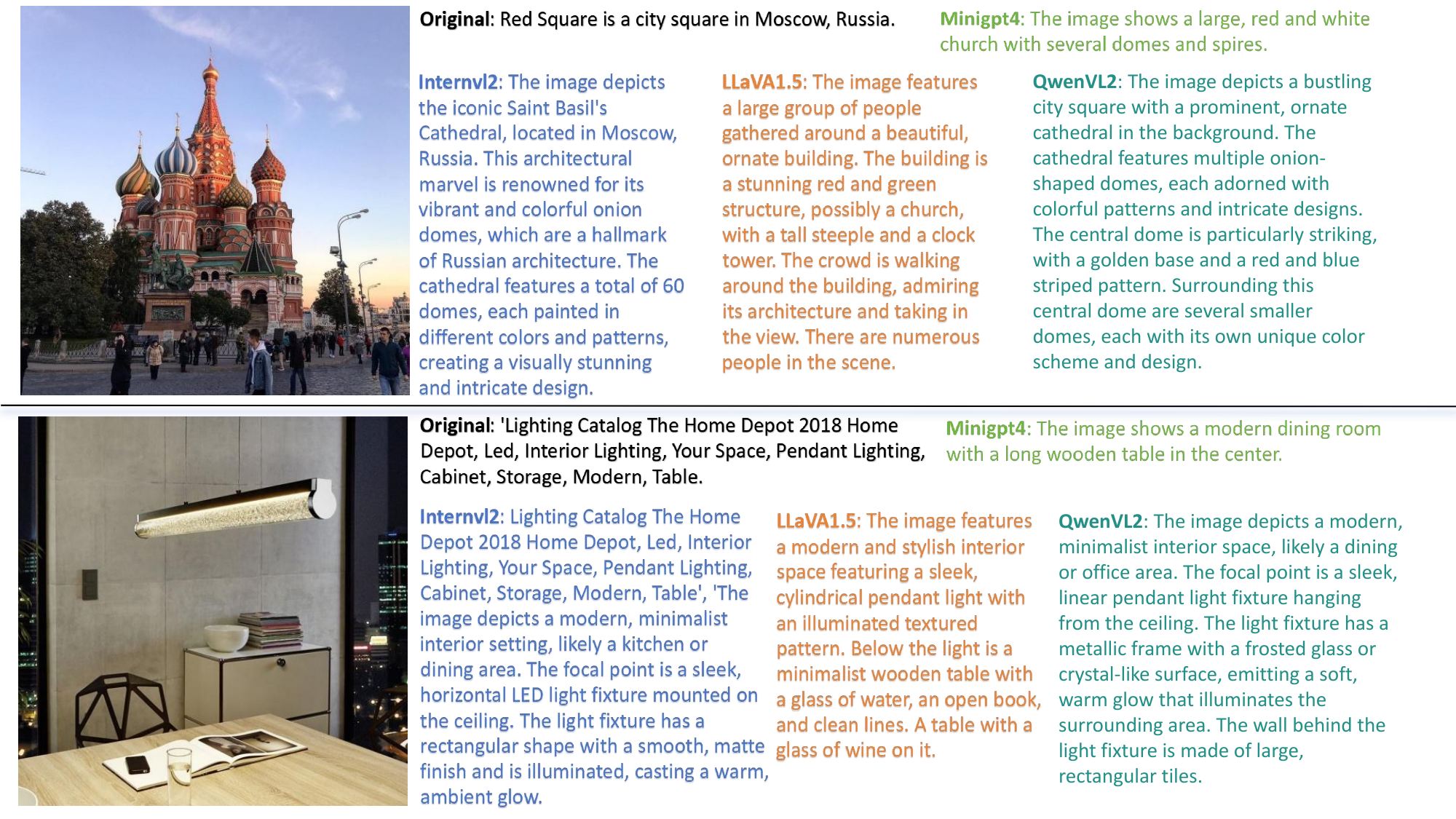}
\end{center}
\vspace{-0.6cm}
\caption{\textbf{Examples of (Image, Multi-Texts) Data from Multi-VLMs.}}
\label{fig:vlm1}
\end{figure*}

\begin{figure*}[t]
\begin{center}
\includegraphics[width=0.95\textwidth] {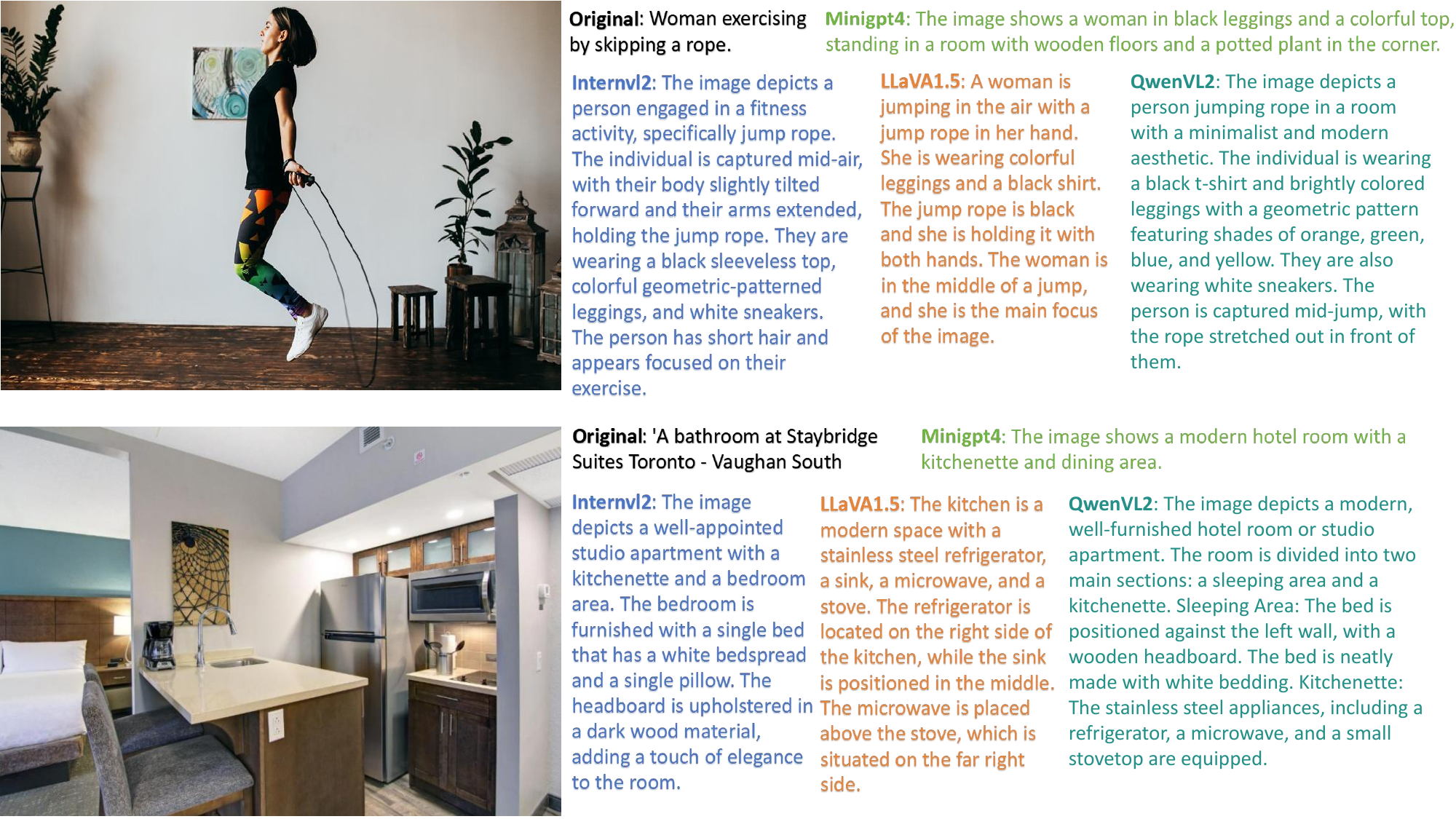}
\end{center}
\vspace{-0.6cm}
\caption{\textbf{Examples of (Image, Multi-Texts) Data from Multi-VLMs.}}
\label{fig:vlm2}
\end{figure*}

For multi-VLMs data, we adopt the ``Details'' prompt in Table~\ref{tab:prompt5} for all VLMs. In Figure~\ref{fig:vlm1} and~\ref{fig:vlm2}, we provide some caption examples. Overall, the texts generated by different VLMs is rather similar. In the main paper, we choose four captions (without style) from multi-prompts to form data for the sake of fairness.

\begin{table}[h]
\centering
\small
\caption{\textbf{Statistics for Multi-VLMs Data.} We calculate the average number of words and the caption variance of multiple VLMs.}
\vspace{-0.25cm}
\begin{tabular}{c|c|c}
\hline
\textbf{VLM} & \textbf{AVG Words} & \textbf{Word Variance} \\ \hline \hline
Original & 17.3 & 163.4 \\ \hline
InternVL2~\cite{chen2024internvl} & 60.8 & 14.9 \\ \hline
Minigpt4~\cite{zhu2023minigpt} & 16.5 & 24.8 \\ \hline
LLaVA1.5~\cite{liu2024improved} & 48.5 & 286.4 \\ \hline
QwenVL2~\cite{bai2023qwen} & 58.5 & 23.6 \\ \hline
\end{tabular}
\label{tab:vlms}
\end{table}

\subsection{Dataset Statistics}
Table~\ref{tab:vlms} and~\ref{tab:prompts} report statistics by adopting different VLMs and prompts. The generated captions have different length distributions, forming long/short counterparts. More recent VLMs like InternVL and QwenVL offer long captions steadily, while Minigpt4 tends to give a short description. For multi-prompts data, captions for main object and background descriptions vary much in length, highly depending on the image semantics (object oriented or complex scene).

\begin{table}[h]
\centering
\small
\caption{\textbf{Statistics for Multi-Prompts Data.} We calculate the caption variance of InternVL2-8B~\cite{chen2024internvl} using different prompts.}
\vspace{-0.25cm}
\begin{tabular}{c|c|c}
\hline
\textbf{Prompt} & \textbf{AVG Words} & \textbf{Word Variance} \\ \hline \hline
Details & 60.8 & 14.9 \\ \hline
Nouns & 10.7 & 61.8 \\ \hline
Main Object & 31.7 & 382.0 \\ \hline
Background & 42.3 & 432.0 \\ \hline
Style & 56.2 & 53.3 \\ \hline
\end{tabular}
\label{tab:prompts}
\end{table}

\subsection{Caption Details}
For multi-VLM captioning, we utilize LLaVA1.5-13B~\cite{liu2024improved}, Minigpt4-7B~\cite{zhu2023minigpt}, QwenVL2-7B~\cite{bai2023qwen}, as well as InternVL2-8B~\cite{chen2024internvl}. For multi-prompt captioning, we utilize InternVL2-8B~\cite{chen2024internvl} for its superior instruction-following performance. We set the max number of output tokens to be 77 due to the length limitation of CLIP~\cite{clip}. On CC12M~\cite{changpinyo2021conceptual} dataset, it takes around 2000-2400 A100-80G GPU hours (depending on the used VLM and prompt) to generate one typical caption (500-700 GPU hours on CC3M~\cite{changpinyo2021conceptual}). After captioning, we set a minimum length (string length$>$10) and threshold on clip score to filter out bad cases, then we re-caption those unqualified data until they meet the standard.

\section{Visualizations}
\begin{figure*}[t]
\begin{center}
\includegraphics[width=0.7\textwidth] {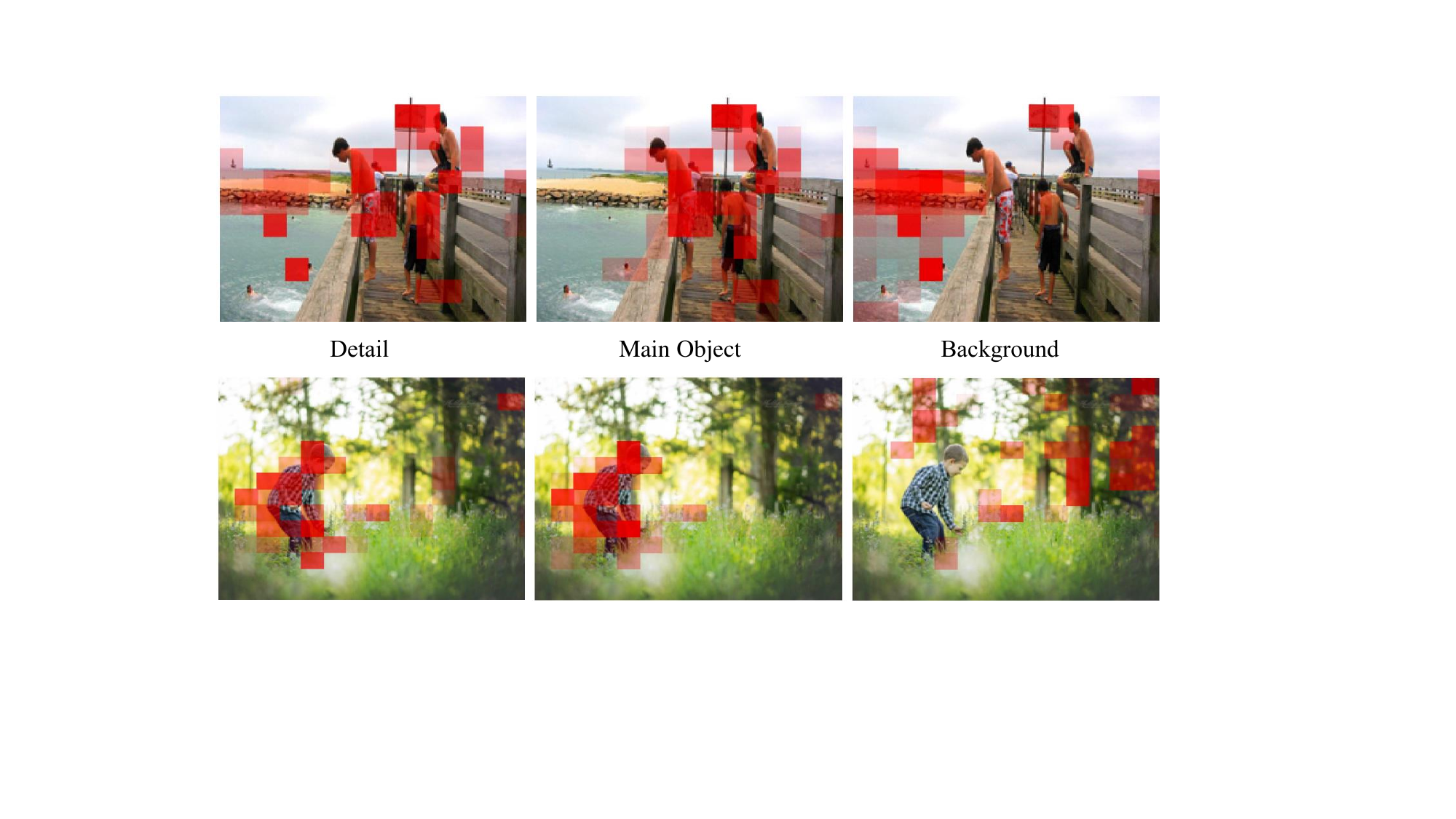}
\end{center}
\vspace{-0.5cm}
\caption{\textbf{Visualization of Attention Maps Among Different Visual Class Tokens.}}
\label{fig:vis1}
\end{figure*}

Here we offer more visualizations on the attention distribution among different image class tokens in $\Psi_{\mathrm{MLP}}$. In Figure~\ref{fig:vis1}, we demonstrate that our M2M contrastive learning paradigm allows to produce image features that summarize visual semantics from various aspects.

\section{Limitations \& Future Work}
Due to limited computing, we do not fully explore the case where $M$ is large (multi-VLMs and multi-prompts) or conduct experiments on larger datasets ({\em e.g.} Laion400M~\cite{schuhmann2021laion}). Nevertheless, our exploration on holistic alignment is anticipated to bring significant benefits for the VLM domain. In reality, there exist naturally occurring datasets comprising (image, multi-text) pairs, such as those found in commodity scenario (commodity attribute/merchant description/buyer comments), where our idea can show the capability.

\end{document}